\documentclass[sn-mathphys,Numbered]{sn-jnl}

\usepackage{graphicx}%
\usepackage{multirow}%
\usepackage{amsmath,amssymb,amsfonts}%
\usepackage{amsthm}%
\usepackage{mathrsfs}%
\usepackage[title]{appendix}%
\usepackage{xcolor}%
\usepackage{textcomp}%
\usepackage{manyfoot}%
\usepackage{booktabs}%
\usepackage{algorithm}%
\usepackage{algorithmicx}%
\usepackage{algpseudocode}%
\usepackage{listings}%
\usepackage{times}
\usepackage{epsfig}
\usepackage{graphicx}
\usepackage{subcaption}
\usepackage{amsmath}
\usepackage{amssymb}
\usepackage{multirow}
\usepackage{diagbox}
\usepackage{colortbl}
\usepackage{booktabs}
\usepackage{pifont}
\usepackage{makecell}

\usepackage{graphicx}
\usepackage{amsmath}
\usepackage{amssymb}
\usepackage{booktabs}
\usepackage{enumitem}
\usepackage{multirow}
\usepackage{diagbox}
\usepackage[figuresright]{rotating}
\usepackage{tabularx}
\usepackage{longtable}
\usepackage{makecell}
\usepackage{xcolor}
\usepackage{lscape}
\usepackage{adjustbox}

\begin{document}

\title[Article Title]{Cross-domain Few-shot Object Detection with Multi-modal Textual Enrichment}


\author[1]{\fnm{Zeyu} \sur{Shangguan}}\email{zshanggu@usc.edu}

 \author[1]{\fnm{Daniel} \sur{Seita}}\email{seita@usc.edu}

\author[1]{\fnm{Mohammad} \sur{Rostami}}\email{rostamim@usc.edu}

\affil*[1]{\orgdiv{Department of Computer Science}, \orgname{University of Southern California}, \orgaddress{ \city{Los Angeles},  \state{CA}, \country{USA}}}


\abstract{Advancements in cross-modal feature extraction and integration have significantly enhanced performance in few-shot learning tasks. However, current multi-modal object detection (MM-OD) methods often experience notable performance degradation when encountering substantial domain shifts.
We propose that incorporating rich textual information can enable the model to establish a more robust knowledge relationship between visual instances and their corresponding language descriptions, thereby mitigating the challenges of domain shift. Specifically, we focus on the problem of Cross-Domain Multi-Modal Few-Shot Object Detection (CDMM-FSOD) and introduce a meta-learning-based framework designed to leverage rich textual semantics as an auxiliary modality to achieve effective domain adaptation.
Our new architecture incorporates two key components:
(i) A multi-modal feature aggregation module, which aligns visual and linguistic feature embeddings to ensure cohesive integration across modalities.
(ii) A rich text semantic rectification module, which employs bidirectional text feature generation to refine multi-modal feature alignment, thereby enhancing understanding of language and its application in object detection.
We evaluate the proposed method on three cross-domain object detection benchmarks and demonstrate that it significantly surpasses existing few-shot object detection approaches. The implementation of our method is publicly accessible at: https://github.com/zshanggu/HTRPN.\footnote{Partial results of this work is presented at the  2025 Winter Conference on Computer Vision~\cite{shangguan2024cross}.}}

\keywords{few-shot object detection, semi-supervised learning, region proposal network}




\maketitle
\section{Introduction}
\label{sec:intro}
In real-world industrial settings, many existing deep learning-based object detection methods struggle to detect product defects effectively due to the challenge of data annotation~\cite{shangguan2024improved,han2024few,guirguis2024uncertainty}. These challenges primarily stem from two factors: the limited availability of domain-specific training data and the significant domain gap between pre-trained datasets which are typically sourced from everyday scenarios and the specialized domain of product defect detection.
Even for humans, identifying subtle product defects can be challenging without prior visual experience in the specific domain. However, human experts often rely on detailed training manuals containing textual instructions and guidelines to bridge this gap. These manuals provide critical context and descriptive information, enabling experts to identify defects more accurately and consistently annotate images.
Inspired by this process, we propose a novel methodology that integrates multi-modal rich textual information to enhance object detection performance in scenarios characterized by limited training data~\cite{li2022pseco,weng2022autonomous} and out-of-domain generalization~\cite{rostami2021detection,le2021poodle,jeong2020ood} challenges. By mimicking the human approach of combining visual and textual cues, our method seeks to improve the model's ability to generalize to new domains and effectively detect defects in industrial applications.

We explore the general task of few-shot object detection (FSOD), which focuses on training models capable of identifying objects from classes where only a limited number of labeled examples are available~\cite{Wang20TFA,Shangguan23HTRPN}. This capability is particularly valuable in domains where   rare or less common objects is frequent, such as in autonomous driving, where the ability to adapt to novel scenarios is critical. The typical FSOD workflow involves an initial pre-training phase on base classes with abundant annotated data, followed by fine-tuning on novel classes that have only a few labeled samples. In traditional FSOD settings, base and novel classes generally belong to the same feature domain, which simplifies adaptation but does not always reflect real-world challenges.
Current FSOD methods primarily adopt one of two approaches: fine-tuning or meta-learning~\cite{Wang20TFA}. Fine-tuning, while computationally efficient and requiring fewer GPU resources, often exhibits limited generalization capabilities, especially when applied to novel domains. On the other hand, meta-learning methods achieve stronger generalization due to their class-agnostic training paradigm, enabling them to infer novel classes more accurately. However, this robustness comes at the cost of higher computational demands, as these methods typically require substantial GPU resources~\cite{Zhang23MetaDETR}.

\begin{figure}[t]
    \centering
    \includegraphics[width=\linewidth]{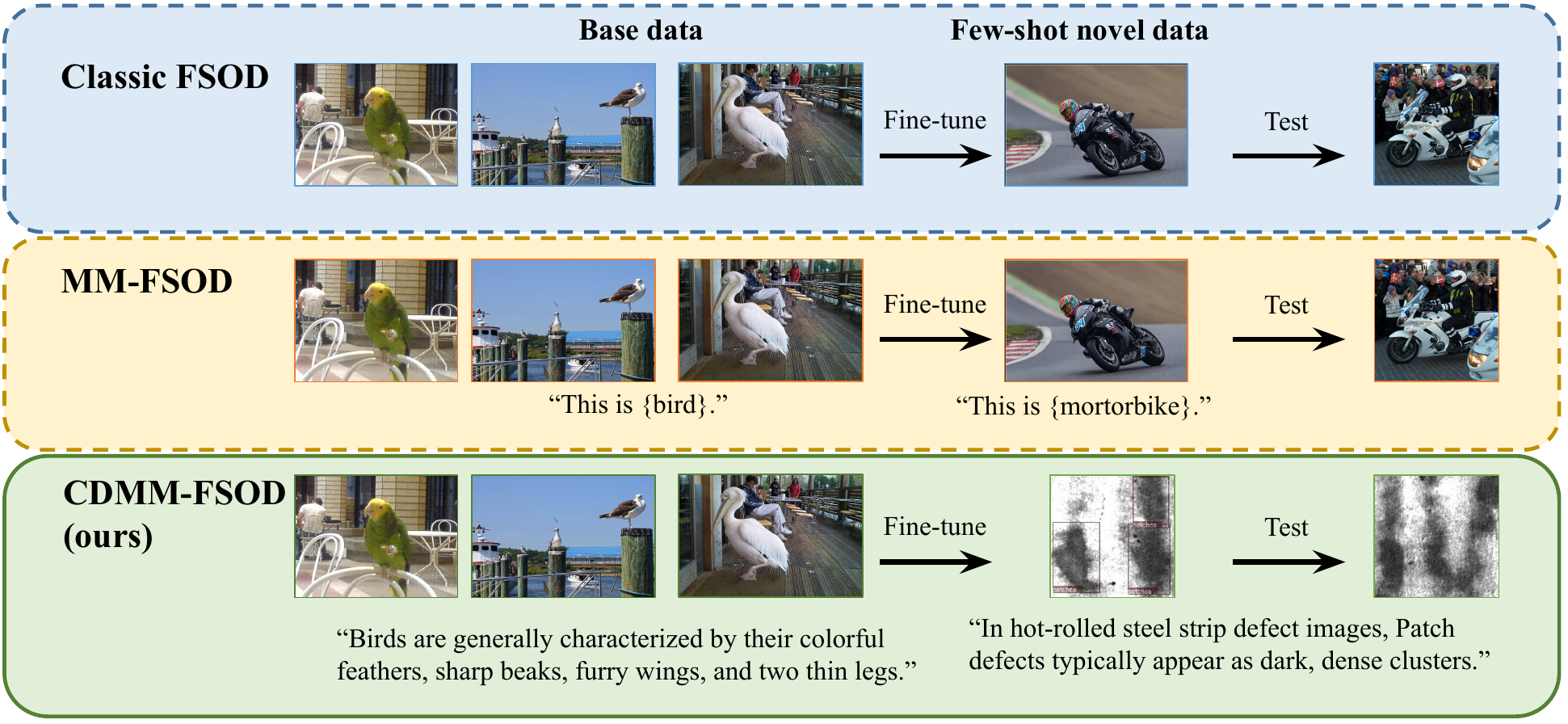}
    \caption{Different FSOD tasks: The classic FSOD task (top) relies solely on visual information for object detection. MM-FSOD (middle) enhances FSOD performance by incorporating a language modality, providing additional contextual information. Building on this approach, our proposed CDMM-FSOD task (bottom) is specifically designed for cross-domain scenarios, extending MM-FSOD by utilizing richer, more detailed text descriptions. The example demonstrates a cross-domain challenge: the model is trained on images and text of common objects (e.g., birds) but must generalize to detect significantly different and less common objects (e.g., patch defects).}
    \label{fig:intro}
\end{figure}

A promising new direction in FSOD is multi-modal learning, which introduces additional modalities such as text to enrich visual feature representations. This approach not only improves performance but also opens the door to zero-shot learning~\cite{bansal2018zero,kolouri2018joint,chao2016empirical}, where models can recognize novel classes without direct training examples (see Figure~\ref{fig:intro}). Multi-modal FSOD encompasses a range of techniques, from straightforward visual question-answering frameworks~\cite{cai2024clumo,Han2024FMFSOD} to more advanced systems that tightly integrate vision and text modalities~\cite{han2023multimodal,azeem2024unified}. The broader field of multi-modal machine learning leverages diverse data sources, including vision, language, acoustics, and tactile inputs, to enhance performance across tasks. This paradigm has demonstrated remarkable success in areas such as visual question-answering~\cite{driess2023palme,srinivasan2022climb}, robotics~\cite{rt22023arxiv,vision_touch_2019}, continual learning~\cite{cai2023task}, and medical image analysis~\cite{Kline2022MultimodalML}.
Multi-modal learning is also foundational to modern generative AI systems like GPT-4~\cite{GPT-4} and Gemini~\cite{Gemini}, which can seamlessly process and produce text and images to perform a wide range of complex tasks.

\begin{figure}[t]
    \centering
    \includegraphics[width=\linewidth]{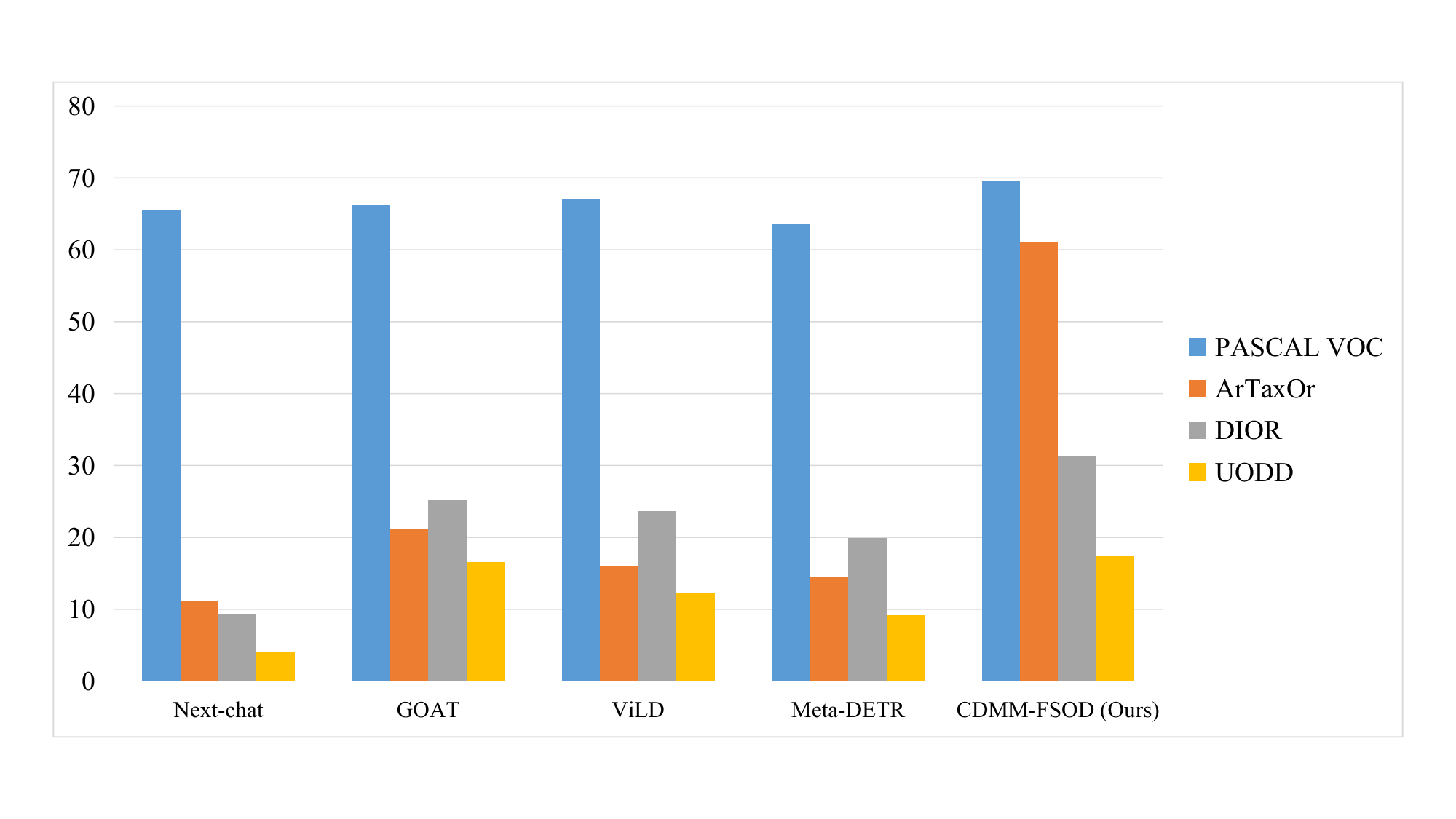}
    \caption{
    Performance results on 10-shot object detection on multiple cross-domain, few-shot datasets: We observe substantial cross-domain degradation for existing MM-OD Next-chat~\cite{zhang2023nextchat}, GOAT~\cite{Wang23GOAT}, and ViLD~\cite{gu2022openvocabulary}, as well as for a single-modal detection method, Meta-DETR~\cite{Zhang23MetaDETR}. In contrast, our proposed method has stronger performance on out-of-domain data. }
    
    \label{fig:degradation}
\end{figure}

While multi-modal object detection (MM-OD) methods have shown considerable success in few-shot learning tasks, their performance often deteriorates in practical scenarios where there is a significant domain gap between the source and target data~\cite{rostami2023domain,zhao2023dual,fu2023styleadv}.   Domain gap presents a critical challenge, as the data distributions in the training (source) and testing (target) domains can differ substantially in real-world detection problems. Such cross-domain, data-scarce scenarios exacerbate the difficulty of maintaining high performance. As illustrated in Figure~\ref{fig:degradation}, object detection methods, including MM-OD approaches, experience significant performance degradation when trained and evaluated on datasets with distinct source and target domains.
To address this issue, we focus on enhancing the robustness of few-shot object detection (FSOD) models by tackling the domain adaptation problem inherent in these scenarios. Specifically, our objective is to address the task of cross-domain multi-modal few-shot object detection (CDMM-FSOD), which aims to bridge the domain gap effectively while facilitating the transfer of information from the source domain to the target domain. By developing strategies to mitigate this domain shift, we seek to improve the generalization and adaptability of MM-OD methods, enabling them to perform reliably in diverse and challenging real-world environments.

More specifically, we hypothesize that leveraging rich, detailed text within a sentence can be significantly more effective than relying solely on simple semantic descriptions, particularly when dealing with target domain images that require technical language for accurate depiction. This hypothesis holds special relevance in scenarios where most models are pre-trained on datasets reflecting common, everyday contexts, leaving them ill-equipped to handle domain-specific technical descriptions. Despite this potential, a key challenge lies in determining how neural network models can effectively utilize such rich text information for detection tasks, as this remains an underexplored area in existing research.
We aim to explore whether and how rich textual information can help address the cross-domain data gaps that hinder object detection performance. As illustrated in Figure~\ref{fig:intro}, our proposed approach for cross-domain multi-modal few-shot object detection (CDMM-FSOD) differs fundamentally from traditional FSOD and multi-modal FSOD (MM-FSOD) methods. Our methodology incorporates rich text descriptions containing detailed technical information about the training data. In contrast, the novel data, such as the patch defect depicted in the example, represents a substantial domain shift. This scenario, where the target domain diverges significantly from the source domain, is not only theoretical but also prevalent in industrial applications. We believe our approach provides a practical and innovative solution to these challenges, addressing a critical gap in the ability of models to generalize effectively across domains in real-world settings.

We adopt Meta-DETR~\cite{Zhang23MetaDETR}, a meta-learning variant of the Detection Transformer (DETR)~\cite{Carion20DETR}, as our baseline due to its Transformer-based architecture and strong capability for adapting to multiple modalities. Unlike traditional multi-modal approaches, our framework incorporates a rich text description for each training category, which can include technical terminology specific to the domain. These rich text descriptions serve as a linguistic support set during the meta-learning process, functioning in a manner analogous to the image-based support set but providing complementary information through language. 
To effectively integrate vision and language information, we design a \textbf{meta-learning multi-modal aggregated feature module}, which fuses the visual and textual embeddings and projects them into a shared class-agnostic feature space. This meta-support feature serves as the foundation for learning across domains. Additionally, to ensure the model captures detailed knowledge embedded in the rich semantics of the text, we introduce a \textbf{rich semantic rectify module}. This module aligns the generated language embeddings with the corresponding ground-truth embeddings, reinforcing the semantic consistency and enhancing the model's understanding of technical language.
Through extensive experiments, we demonstrate that incorporating rich text significantly improves performance, showcasing the effectiveness of our approach in leveraging linguistic information. Moreover, we investigate the impact of text length on performance, providing insights into how the richness and detail of text descriptions influence the model's ability to generalize across tasks.
Our specific contributions include:
\begin{itemize}[noitemsep,leftmargin=*]
    \item We propose a meta-learning multi-modal aggregated feature module that utilizes rich text semantics to alleviate the performance degradation due to domain gaps.
    \item We propose a rich semantic rectify module that learns to align the generated language embedding and the ground truth language embedding, enabling the model to better understand the text for knowledge transfer.
    \item Our experiments on major cross-domain detection data demonstrate that our approach outperforms existing methods, suggesting that multi-modal learning with rich text is effective for detection of few-shot, out-of-domain data.
\end{itemize}

\section{Related Works}
\label{Related Works}

Our work adopts the notion of cross-domain knowledge transfer in the context of few-shot object detection. We review the related works in this research areas.

\subsection{Few-shot Object Detection}
\label{sec:FSOD}

Two prominent training paradigms dominate the FSOD methods: fine-tuning-based methods and meta-learning-based approaches. Fine-tuning methods typically involve a two-stage process. First, a model is pre-trained on a base dataset where ample labeled samples are available for each category. Following this, the model is fine-tuned directly on the novel categories, leveraging the few available labeled samples to adapt its detection capabilities. Notable works in this domain include TFA~\cite{Wang20TFA}, FSCE~\cite{Sun21FSCE}, FSRC~\cite{Shangguan23FSRC}, and DeFRCN~\cite{Qiao21DeFRCN}, each of which introduces innovations to improve fine-tuning performance on novel tasks.
Meta-learning methods, often referred to as ``learning to learn,'' take a fundamentally different approach. Instead of focusing on specific categories, these methods aim to train models to acquire a generalizable metric learning ability. The training process is structured into tasks, or episodes, each designed to mimic the few-shot scenario. By employing architectures like Siamese networks~\cite{Koch2015SiameseNN}, they measure the similarity between support and query features, enabling classification based on learned, class-agnostic knowledge~\cite{Chen21MetaBaseline}. During training, the model is exposed to episodic tasks comprising a support set and a query set. The support set typically represents an \(n\)-way \(k\)-shot setup, where \(n\) categories are included, each with \(k\) labeled examples. These tasks enable the model to generalize across a wide range of unseen categories. 
A few examples of meta-learning-based FSOD methods include Meta-FRCN~\cite{han2022meta} and Meta-DETR~\cite{Zhang23MetaDETR}, which leverage this episodic framework to enhance detection performance on novel categories.  

A FSOD is addressing catastrophic forgetting, where models lose their ability to recognize base categories when fine-tuned on novel categories. To mitigate this issue, researchers have explored strategies that incorporate both base and novel data during training. This balanced approach ensures that the model retains its performance on base categories while adapting to novel ones.
Fine-tuning and meta-learning methods adopt distinct strategies for handling this challenge. Fine-tuning methods generally focus on directly transferring the model’s learning to novel data in a class-specific manner. These methods leverage the pre-trained model’s knowledge and adapt it to new categories through targeted adjustments. Conversely, meta-learning approaches take a broader perspective, aiming to produce class-agnostic models. By learning a generalizable metric scheme that measures the similarity between query and support features, meta-learning enables models to adapt to novel categories without requiring class-specific knowledge.
Meta-learning methods offer a notable advantage in their ability to rapidly adapt to new tasks, attributed to their task-level feature representation. This adaptability often results in higher precision when confronted with diverse and unfamiliar scenarios. However, fine-tuning methods are not without merit. For instance, TFA~\cite{Wang20TFA} demonstrates that fine-tuning methods have the potential to surpass meta-learning approaches in detection performance under certain conditions. On the other hand, studies such as FCT~\cite{Han22FCT} highlight the enduring strengths of meta-learning, emphasizing its power in FSOD scenarios and its superior capability for rapid task adaptation due to its reliance on task-level feature representations.
These contrasting characteristics illustrate the trade-offs between fine-tuning and meta-learning in FSOD. While fine-tuning excels in leveraging pre-trained knowledge for specific categories, meta-learning shines in its flexibility and robustness in rapidly adapting to novel tasks. Both paradigms continue to shape the evolving landscape of FSOD research.

In FSOD research, traditional backbone networks have played a foundational role in driving progress. Well-established architectures such as Faster R-CNN~\cite{RenHGS15}, YOLO~\cite{redmon2016yolo}, and Vision Transformers (ViT)~\cite{dosovitskiy2020vit} are widely utilized due to their proven effectiveness in various detection tasks. These models have provided strong baselines for tackling FSOD challenges, offering robust feature extraction and prediction capabilities.
In recent years, the Detection Transformer (DETR)~\cite{Carion20DETR} has emerged as a compelling alternative for the traditional backbones, gaining traction for its impressive performance compared to classical frameworks. Unlike traditional architectures, DETR leverages a transformer-based design that excels at modeling global context and long-range dependencies, making it particularly advantageous for complex detection scenarios. Additionally, its transformer foundation seamlessly supports integration with natural language processing (NLP) tasks, enabling multi-modal capabilities that align with evolving FSOD requirements.
Given DETR's advantages, our work adopts the meta-learning framework provided by Meta-DETR~\cite{Zhang23MetaDETR}. Our approach combines the strengths of the transformer-based architecture with the adaptability of meta-learning paradigms, allowing us to effectively address the challenges of few-shot detection.  

\subsection{Prompt Learning and Multi-modal Few-shot Object Detection}
\label{sec:MM-FSOD}

Multi-modal object detection (MM-OD) expands the conventional object detection paradigm by integrating additional sources of information beyond visual data, enabling more nuanced and robust detection capabilities. In our approach, language serves as the supplementary modality, transforming the task into training a model on paired image-text data. This integration allows the model to leverage linguistic context alongside visual cues, opening new avenues for detecting and understanding objects in diverse scenarios.
A common strategy in MM-OD networks involves utilizing pre-trained vision-language representations, with CLIP~\cite{Radford2021CLIP} being a prominent choice. These representations, derived from extensive training on paired image-text data, are particularly effective for tasks such as open-vocabulary detection (OVD). In OVD, the model is expected to recognize novel object classes solely through image-text pairs, even without explicit class-specific training. This ability to generalize beyond the training set is a hallmark of multi-modal representations. However, current approaches often rely on relatively simple language inputs, such as class names or basic attributes~\cite{gu2022openvocabulary}, which may limit the richness of the model's understanding.
The advent of large language models (LLMs) has further accelerated advancements in MM-OD, enabling more sophisticated interactions between vision and language. For instance, Next-Chat~\cite{zhang2023nextchat} introduces a Q\&A-based framework that combines object detection with instance segmentation, allowing the model to interactively answer questions about detected objects. Similarly, YOLO-World~\cite{cheng2024yoloworld} employs a region-text contrastive learning approach to build a visual-linguistic interactive model, emphasizing the dynamic interplay between textual prompts and localized visual regions.

Multi-modal few-shot object detection (MM-FSOD) remains a relatively underexplored area within the broader domain of object detection. The task extends the principles of few-shot learning to multi-modal contexts, aiming to train models capable of detecting objects with minimal examples while leveraging both visual and linguistic information. Recent efforts, such as the framework proposed by Han et al.~\cite{han2023multimodal}, introduce innovative approaches to this problem. Their method incorporates soft prompt tokens to achieve low-shot object detection by embedding language cues directly into the learning process. 
However, current methods often rely on simplistic text inputs, such as image-level descriptions or category names, as the language label for objects within an image. While effective in some scenarios, this limited textual representation poses challenges when significant domain gaps exist between the base and novel datasets. In such cases, the sparse linguistic information fails to provide the nuanced context needed for effective knowledge transfer, potentially hindering the model’s ability to generalize to novel categories.
To address these limitations, our work builds upon the foundational scheme of incorporating linguistic information into MM-FSOD frameworks. We propose using richer and more semantically complex language descriptions, which provide deeper contextual understanding and better alignment between visual and textual modalities. By employing diverse and detailed textual inputs, our multi-modal network is designed to bridge the domain gap more effectively, enabling superior generalization from base classes to novel datasets. Our approach represents a significant step forward in leveraging the synergy between vision and language for few-shot object detection, particularly in scenarios with challenging domain shifts.

\subsection{Cross-domain Few-shot Object Detection}
Cross-domain few-shot object detection (CD-FSOD) addresses the challenges of performing FSOD when the source and target feature domains exhibit significant differences~\cite{rostami2019sar,fu2025cross}. This scenario is particularly relevant in real-world applications where models trained on one domain must adapt to detect objects in a vastly different domain. For example, a model pre-trained on natural images may need to detect objects in medical images or aerial photographs.
To simulate such cross-domain scenarios, MoFSOD~\cite{Lee22Rethink} introduces a multi-domain FSOD benchmark comprising 10 datasets from diverse domains, offering a comprehensive framework to evaluate CD-FSOD performance. Their approach defines \( k \)-shot sampling as \( k \) images per class and highlights the critical role of leveraging pre-training datasets effectively to boost model performance. By focusing on diverse domains, MoFSOD provides insights into the challenges of domain adaptation in FSOD.
Expanding on this approach, CD-FSOD~\cite{Xiong2023CDFSOD} presents a cross-domain FSOD benchmark with balanced data per category, where each category is represented by \( k \) instances. This design aligns more closely with standard \( k \)-shot FSOD benchmark protocols, such as those proposed in~\cite{Wang20TFA}. Additionally, CD-FSOD introduces a novel distillation-based baseline, further advancing the field by addressing the knowledge transfer challenges between domains.
Acrofod~\cite{Gao22CDFSOD} takes a different approach by designing an adaptive optimization strategy to select appropriate augmented data for cross-domain training. This method enhances the model's ability to generalize across domains by strategically leveraging data augmentations.

Building upon these prior works, our research seeks to mitigate the degradation often observed in CD-FSOD performance by utilizing multi-modal feature alignment. By integrating complementary information from multiple modalities, we aim to enhance the model's capacity to bridge domain gaps. Our evaluation primarily follows the benchmark provided by Acrofod~\cite{Gao22CDFSOD}, ensuring that our approach is rigorously tested against established cross-domain FSOD scenarios. This combination of multi-modal alignment and robust benchmarking represents a promising step toward addressing the unique challenges of CD-FSOD.

\section{Proposed Methods}
\label{sec:Proposed Methods}

\subsection{Preliminaries}
\label{sec:Preliminaries}

In the conventional FSOD paradigm, the label set is divided into two distinct groups: \emph{base} classes, denoted as \( C_B \), and \emph{novel} classes, represented as \( C_N \). Classes in \( C_B \) have access to a substantial amount of labeled training data, whereas each class in \( C_N \) is characterized by a limited number of labeled instances. Importantly, it is assumed that the sets \( C_B \) and \( C_N \) are disjoint, meaning there is no overlap between the base and novel class categories. 
The training dataset, \(\mathcal{D}_{\rm FSOD} = \{(x, y)_i\}_{i=1}^{N_D}\), comprises \(N_D\) annotated samples. Each sample consists of an RGB image \(x\) and its corresponding ground-truth annotations \(y\). The annotation \(y\) is structured as \(y = \{(c_j, b_j) \mid j \in \{1, 2, \ldots, N_{\rm obj}\}\}\), where \(N_{\rm obj}\) represents the number of objects present in the image \(x\). For each object \(j\), \(c_j \in C_B \cup C_N\) specifies its class label, and \(b_j\) defines its bounding box coordinates.
The primary goal of FSOD is to train a model on \(\mathcal{D}_{\rm FSOD}\) such that it performs well at test time, particularly in detecting objects that belong to the novel classes \(C_N\), despite the limited labeled instances available for those classes. This challenging task has been addressed in several notable works, such as those by Kang et al.~\cite{Kang19fsodrewei} and Wang et al.~\cite{Wang20TFA}, which explore methods to improve detection performance under the few-shot regime.

The Multi-Modal Few-Shot Object Detection (MM-FSOD) task builds upon the foundational FSOD framework by incorporating additional textual information to enhance the detection process. Specifically, the dataset for this task, denoted as \(\mathcal{D}_\text{MM-FSOD} = \{(x, \ell, y)_i\}_{i=1}^{N_D}\), now includes, for each data point, an RGB image \(x\), ground-truth annotations \(y\), and a set of ``category-specific text labels'' \(\ell\). The text labels \(\ell = \{l_j \mid j \in \{1, 2, \ldots, N_{\rm obj}\}\}\) provide descriptive textual information for each of the \(N_{\rm obj}\) object instances within the image \(x\). 
The textual descriptions \(l_j\) can take various forms, commonly as the ``category name'' or as a ``template-based sentence'' embedding the category name. For example, a typical template might be: \texttt{This is a picture of <category>}. Beyond manually defined templates, text descriptions can also be generated automatically using external image-text models like CLIP~\cite{Radford2021CLIP}, which maps image content to meaningful text descriptions. Such automated approaches enable generating diverse and contextually relevant textual annotations for objects in the dataset, as explored in recent works~\cite{han2023multimodal}.
We extend MM-FSOD to Cross-Domain Multi-Modal Few-Shot Object Detection (CDMM-FSOD). In this setting, the distinction between base classes \(C_B\) and novel classes \(C_N\) becomes more pronounced, with classes in \(C_B\) and \(C_N\) belonging to substantially different domains. This cross-domain scenario introduces additional challenges, as the model must not only handle the limited samples in \(C_N\) but also generalize effectively across domain gaps between \(C_B\) and \(C_N\), leveraging both image and textual information to bridge these differences.

\subsection{Rich Text}
\label{sec:Rich Text}

We denote the text data corresponding to the base and novel categories as \( T_B \) and \( T_N \), respectively. To construct the textual input \(\ell\) for the training data, we manually create detailed and fixed text descriptions for each category \( c \in C_B \cup C_N \). These descriptions form a set denoted as \( W^C = \{w^1, w^2, \ldots, w^C\} \), where \( C = |C_B \cup C_N| \) represents the total number of categories.
Each text description \( w^i \) is composed of multiple tokens with a variable length \( M \). Formally, \( w^i = \{ w^i_1, w^i_2, \ldots, w^i_M\} \), where the first token \( w^i_1 = \langle s \rangle \) signifies the start of the sentence, and the last token \( w^i_M = \langle /s \rangle \) marks the end of the sentence.
To construct the set \( W^C \), we employ two complementary strategies:
 (i) Aspect-based Description: For each category, we refer to relevant content on Wikipedia and create descriptions that detail aspects such as color, shape, attributes, material, and other intrinsic properties. For instance, a description for ``motorbikes'' might include its structural and functional features.
(ii) Contextual Extension: Beyond the aspects described in the first strategy, we extend the sentences by incorporating contextual details to reflect common visual relationships between categories. For example, the original sentence, ``Motorbikes have two wheels, a motor, and a sturdy frame,'' is expanded to include contextual relationships: ``Motorbikes have two wheels, a motor, and a sturdy frame, usually ridden by a person.'' This enhancement introduces a typical visual interaction between motorbikes and persons, enriching the textual representation for cross-modal understanding.
The complete set of text descriptions, covering all base and novel classes, is provided in the experimental section for each dataset that we have used.

\begin{figure*}[t]
    \centering
    \includegraphics[width=\linewidth]{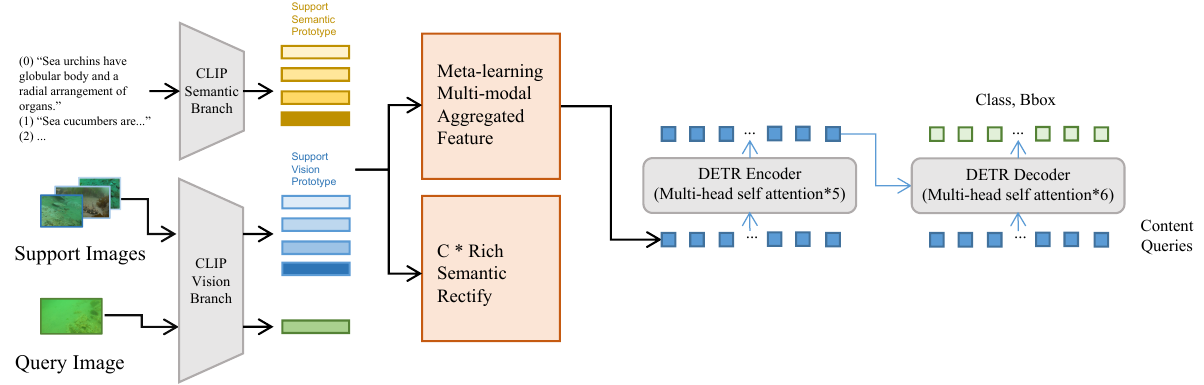}
\caption{The proposed architecture: the proposed ``multi-modal feature aggregation module'' and ``rich text rectification module'' are highlighted in red blocks in Figure~\ref{fig:module}, which provides further details about their design and structure. The ``multi-modal feature aggregation module'' facilitates the fusion of features across different modalities, enabling effective cross-modal embedding integration. Meanwhile, the ``rich text rectification module'' enhances the model's ability to comprehend and leverage information from both image and text modalities. 
We have designed this model to operate in an ``end-to-end manner'', processing a set of support images, query images, and a collection of rich category-specific textual descriptions during training. The model then outputs the detection results for the objects present in the query images, effectively combining visual and textual inputs for improved performance.
}
    \label{fig:pipeline}
\end{figure*}

\subsection{Proposed Architecture}
 
We adopt Meta-DETR~\cite{Zhang23MetaDETR} as the baseline network for our approach. Meta-DETR builds upon Deformable DETR~\cite{zhu2021deformable}, which itself is an advanced object detection model leveraging the transformer architecture~\cite{Transformer2017}. DETR employs a ResNet~\cite{ResNet2016} backbone to extract high-level features from input images. These features are then passed through an encoder that generates memory embeddings. Subsequently, a decoder processes these memory embeddings to produce object proposals, effectively identifying and localizing objects in the image.
In our implementation, we follow Meta-DETR's methodology and utilize a ``weight-shared self-attention module'' to incorporate meta-learning for both the query and support images. This mechanism enables the network to generalize effectively across categories with limited labeled data by learning shared representations.
The classic DETR architecture is composed of both an encoder and a decoder, each consisting of six multi-head attention layers. However, Meta-DETR introduces the meta-learning mechanism selectively, applying it only at the first multi-head attention layer of the encoder. This focused design enhances the network's ability to perform few-shot object detection while maintaining computational efficiency. By leveraging this architecture, our model can effectively integrate query-support relationships and detect objects with minimal data for novel categories.

The meta-learning module begins by processing both query and support images through a shared ResNet feature extractor, yielding their respective basic feature representations. These query and support features are then passed into a multi-head attention module, ensuring they are transformed into a shared feature space for effective comparison and integration. 
In this architecture, the support sample features act as the key (\(K\)) and value (\(V\)) matrices, while the query image features serve as the query (\(Q\)) input to a single-head self-attention mechanism. This self-attention mechanism is designed to perform class-specific feature matching, enabling the network to align query and support features based on their category-specific similarities. 
To address the class-agnostic learning task inherent in meta-learning, an additional parallel branch is introduced. This branch maps the class-specific ``support-query'' relationship into a class-agnostic task ``embedding-query'' relationship. In this branch, Meta-DETR employs an independent, trainable class-agnostic feature prototype denoted as \(V'\). The support and query features are still used as the key (\(K'\)) and query (\(Q'\)) matrices, respectively, within this branch. The single-head self-attention module in this branch transforms the original class-specific support features into class-agnostic prototype features. 

During inference, the model exclusively relies on the class-agnostic prototypes as the support features. These aggregated prototypes, representing general task-relevant embeddings, are fed into the input of the DETR network, which subsequently generates the final detection outputs. This dual-branch approach ensures that the model captures both class-specific and class-agnostic relationships, enabling it to perform robust detection, even for novel categories, under few-shot learning settings.

\begin{figure*}[t]
  \centering
  \includegraphics[width=\linewidth]{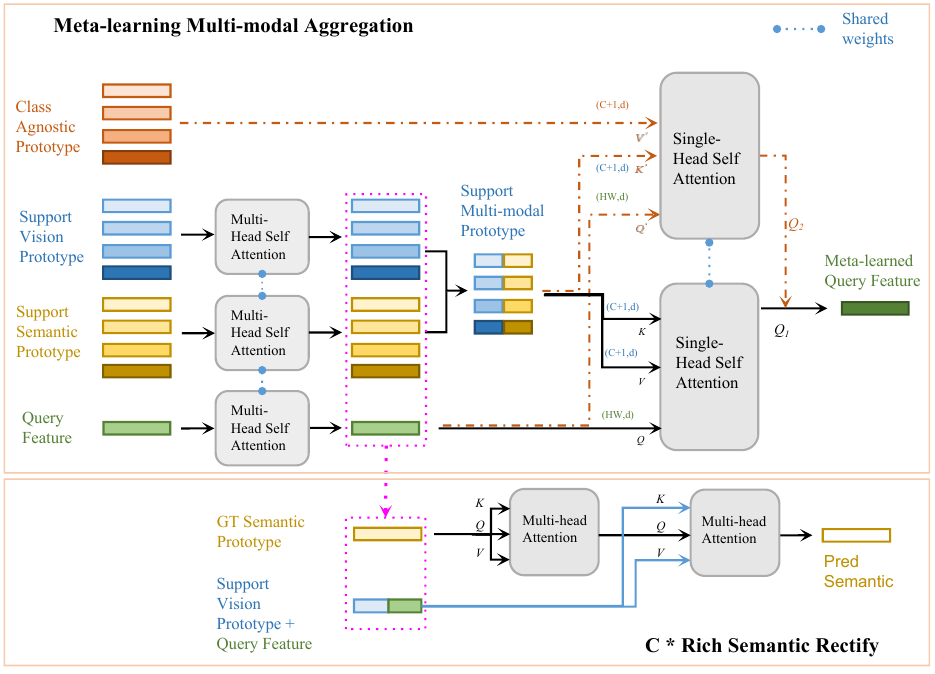}
  \caption{Details of our meta-learning multi-modal aggregation module (\emph{upper region}) and the rich semantic rectify module (\emph{lower region}). Different colors are used for different feature branches. The rich semantic rectify module is  used only during training, and not at test time. 
  }
  \label{fig:module}
\end{figure*}

The overall architecture of our proposed method is illustrated in Figure~\ref{fig:pipeline}. In our design, we incorporate ``rich text labels'' as auxiliary support features to enhance the model's understanding of category-specific information. These text labels are combined with the visual support features, enabling the joint representation of image and language data. The fused support features are then utilized to train the ``single-head self-attention meta-learning module'', referred to as the Meta-Learning Multi-Modal Feature Aggregation Module (MM Aggre.).
Additionally, to ensure the model effectively leverages the semantic information provided by the rich text features, we introduce a dedicated component called the ``Rich Text Semantic Rectification Module (Rich Text Rect.)''. This module performs bidirectional generation of language features, refining the text embeddings to align closely with the ground-truth language features. This alignment process not only improves the model's ability to integrate textual knowledge but also enhances its overall robustness and accuracy in handling multi-modal tasks.

\subsection{Multi-modal Feature Aggregation}

The multi-modal feature aggregation module forms the core of our framework, designed to integrate information from multiple modalities effectively. This module is capable of fusing features across multiple support categories simultaneously, making it a key component in handling the complexities of multi-modal tasks. A detailed visualization of its architecture is provided in Figure~\ref{fig:module}.
To extract the foundational features, we utilize the pre=trained CLIP model, which generates the primary image features for the support and query sets, as well as the primary semantic features for the rich text labels. For the support image features, a RoIAlign layer followed by an average pooling layer is applied to derive instance-level category prototypes. This operation is essential in object detection tasks because raw image features cannot accurately represent individual object instances, especially in images containing multiple foreground objects.
Additionally, consistent with conventional object detection methods, the background is treated as an extra hidden category and is initialized as one of the category prototypes. This ensures that the model accounts for non-object regions during feature aggregation.
The extracted features from both modalities are then projected into a shared feature space using a multi-head self-attention module, implemented as the first layer of the DETR encoder. This shared-weight mechanism ensures alignment between visual and textual representations, enhancing the model's capability to fuse multi-modal data. The importance of this shared-weight design is validated through an ablation study, as discussed in Section~\ref{sec: ablation}.

Next, we combine the image and semantic features by concatenating and averaging them to form the comprehensive support feature. Specifically, the support vision feature is concatenated with the corresponding language feature from the same category, ensuring that both modalities contribute equally to the final representation. This concatenated feature serves as a unified input for subsequent processing.
To further enhance the learning process, we apply a single-head self-attention module. This module serves two primary functions: (i) it maps the support-query feature relationships, enabling the model to align the features from the support and query sets based on their class-specific similarities, and (ii) it matches the class-agnostic task embedding-query relationship, aligning the task-related features with the query features in a generalizable manner. This dual functionality allows the model to learn both category-specific and class-agnostic representations, optimizing the feature matching process across different categories and tasks.

For the support-query feature mapping, the combined multi-modal feature is used as the key ($K$) and value ($V$) matrices in the single-head self-attention module. Specifically, given $C$ support categories, the multi-modal support feature ($S \in \mathbb{R}^{(C+1) \times d}$) and the query image feature ($Q \in \mathbb{R}^{HW \times d}$) are utilized to compute the feature mapping coefficients, as follows:
\begin{equation}
    A = \text{softmax}\left(\frac{QS^T}{\sqrt{d}}\right)
\label{eq:feature mapping}
\end{equation}

In a query image, only specific foreground objects will have matching features with the support instances, while the rest of the image (i.e., the background) should be disregarded. Therefore, we focus only on the relevant areas of the query feature map and reduce the influence of regions that do not match any support prototype. To achieve this, we apply a sigmoid function to the support prototype, $\sigma(S)$, treating it as the value matrix, and combine it with the coefficient $A$ to form the attention mechanism. 
As a result, the output of the support-query feature mapping process is:
\begin{equation}
    Q_1 = A\sigma(S)\odot Q,
\label{eq:output feature mapping}
\end{equation}
where $\odot$ represents the element-wise Hadamard product. The output $Q_1$ is a refined query feature map that keeps only the `foreground' features.

For the task embedding-query encoding matching, we introduce a set of class-agnostic task prototypes ($T \in \mathbb{R}^{(C+1) \times d}$), which are used to replace the class-specific support prototypes. Essentially, the class-specific support prototypes are mapped to these pre-defined task prototypes, allowing the query feature to align with the task prototypes during the prediction process.
The task prototype $T$ serves as the value matrix ($V'$) for the same single-head self-attention module used in the support-query feature mapping. The initialization of $T$ follows a pattern similar to sinusoidal position embeddings in transformers, which are commonly used to encode positional information in a way that facilitates generalization across different tasks.
The resulting output from this process would be:
\begin{equation}
    Q_2 = AT
\label{eq:output task mapping}
\end{equation}
The overall output of the multi-modal aggregation module is the element-wise addition of $Q_1$ and $Q_2$.

\subsection{Rich Text Semantic Rectify}

The rich text semantic rectify module is illustrated in ~\ref{fig:module}. The core idea behind this module is to employ a transformer encoder architecture~\cite{afham2021rich}, denoted as $f_\theta$, to generate category-level language sequence features for the fused support-query samples. Once these language features are generated, they are expected to align with the corresponding ground-truth language features. This alignment process helps refine the model's understanding and generation of the language representation.
Furthermore, the language generation within this module is bidirectional, meaning that the model generates the sentence both in the forward and backward directions. This bidirectional generation ensures that the model develops a more robust understanding of the image-text relationships and enhances its ability to encode and integrate semantic knowledge from both modalities. By doing so, the module helps improve the model's capacity to bridge the gap between visual and textual information, leading to more accurate and contextually aware predictions.

Specifically, for a given category $c$, its rich text tokens undergo refinement through the multi-head self-attention module that is shared with the vision branch to get its semantic prototype $l^c$, as shown in ~\ref{fig:module}.
Additionally, the support and query features are averaged to create a composite feature $p$, which serves as the $K$ and $V$ matrices for a multi-head self-attention module $A_3$, as shown in ~\ref{fig:module}.
In $A_3$, the refined text tokens act as the $Q$ matrix.
Subsequently, it generates the predicted text sequence $\hat W$ of length $M$.
This entire process is duplicated for both forward and backward prediction components, i.e.,  the model will predict $\hat W$ from both the left-to-right and the right-to-left directions so that it could understand the rich text through its bidirectional context.
The loss function for this purpose would be:
\begin{equation}
\begin{split}
    \mathcal{L}_{rect} = \frac{1}{2} \left[\sum_{i=1}^{C}\sum_{j=2}^{M}-\mathrm{log} f_\theta(\hat w^i_j-w^i_j | p) +
    \sum_{i=1}^{C}\sum_{j=1}^{M-1}-\mathrm{log} f_\theta(\hat w^i_j-w^i_j | p) \right]
\end{split}
\label{eq:rectify loss}
\end{equation}

\section{Experimental Validation}

We validate our method on three suitable benchmarks and compare our results against state-of-the-art methods to demonstrate its effectiveness.

\subsection{Experimental Setup}

We perform experiments using four existing benchmarks. We adopt these datasets such that they can be used in the context of our proposed method.

\subsubsection{Datasets}
\label{sec:datasets}

\paragraph{CD-FSOD}

The CD-FSOD benchmark, introduced by Xiong et al.\cite{Xiong2023CDFSOD}, serves as an exceptional platform for evaluating our experimental framework. This benchmark incorporates diverse datasets, allowing us to assess few-shot learning models in distinct domains. Specifically, we leverage three datasets: ArTaxOr\cite{Drange2020arthropod}, DIOR~\cite{LI2020DIOR}, and UODD~\cite{Jiang2021UODD}, as our target-domain datasets. The ArTaxOr dataset features seven arthropod categories, the DIOR dataset includes 20 classes derived from satellite imagery, and the UODD dataset comprises three underwater species, offering a rich diversity for cross-domain analysis and testing.

\paragraph{CDS-FSOD}

The CDSFSOD dataset, introduced by Fu et al.\cite{fu2025cross}, is constructed by integrating three diverse datasets: Clipart1k\cite{inoue2018cross}, which features clipart-style object images, DeepFish\cite{saleh2020realistic}, containing underwater fish images, and NEU-DET\cite{song2013noise}, designed for detecting surface defects in industrial materials. A balanced few-shot sampling strategy is applied in this dataset, meaning the $k$-shot setting includes exactly $k$ instances for each category across all datasets. This design ensures consistent and equitable representation of classes, making the dataset suitable for cross-domain FSOD evaluations.

\paragraph{PASCAL-VOC}

We conduct evaluations of our method using the widely adopted PASCAL VOC benchmark for general few-shot object detection (FSOD) due to its extensive use in the literature. Following the standard FSOD evaluation protocol proposed by Wang et al.~\cite{Wang20TFA}, 15 PASCAL VOC categories are designated as base classes for pre-training, while the remaining 5 serve as novel classes for fine-tuning in the few-shot setting.  
The general FSOD framework incorporates fine-tuning on both base and novel categories, which effectively mitigates the issue of catastrophic forgetting. To ensure unbiased evaluation, the dataset is divided into three fixed category splits for training and testing, facilitating a robust measurement of the model’s average performance.  
Two distinct sampling strategies are employed for fine-tuning: ``balanced sampling'' and ``unbalanced sampling''. Balanced sampling allocates $k$ instances per category, encompassing both base and novel classes, as seen in methods like TFA. Conversely, unbalanced sampling limits the $k$-shot instances strictly to novel categories, excluding base classes, which is used by methods like Meta-DETR~\cite{Zhang23MetaDETR}. In our experiments, we adopt the unbalanced sampling strategy to focus on novel category performance.

\paragraph{Mo-FSOD}

The MoFSOD dataset, introduced by Lee et al.~\cite{Lee22Rethink}, employs an unbalanced sampling strategy, where the $k$-shot setting refers to $k$ unique images rather than $k$ individual instances, marking a deviation from conventional FSOD approaches. In our experiments, we utilized three balanced-sampling sub-datasets: Clipart1k~\cite{inoue2018cross}, which mirrors PASCAL VOC categories using cartoon-style images; DeepFish~\cite{saleh2020realistic}, consisting of high-resolution images spanning 59 underwater fish species; and NEU-DET~\cite{song2013noise}, focused on detecting surface defects in industrial materials. This combination offers a diverse evaluation framework for FSOD methods.


\subsubsection{Detailed List of the Rich Text}
\label{sec:Detailed List}

We augment the datasets that we use in our experiments with rich texts so we can apply on our algorithm on the resulting datasets. 
Note that the four benchmarks that we use for FSOD evaluation are built using these datasets. To this end, we build two rich texts to describe classes in these datasets. We build rich text manually or by using an LLM~\cite{Sun2021ERNIE3L}. As indicated in our experiments, either way could generate significant performance improvement on the out-of-domain benchmarks.

\paragraph{ArTaxOr Dataset}

\begin{figure*}[t]
    \centering
    \includegraphics[width=0.7\linewidth]{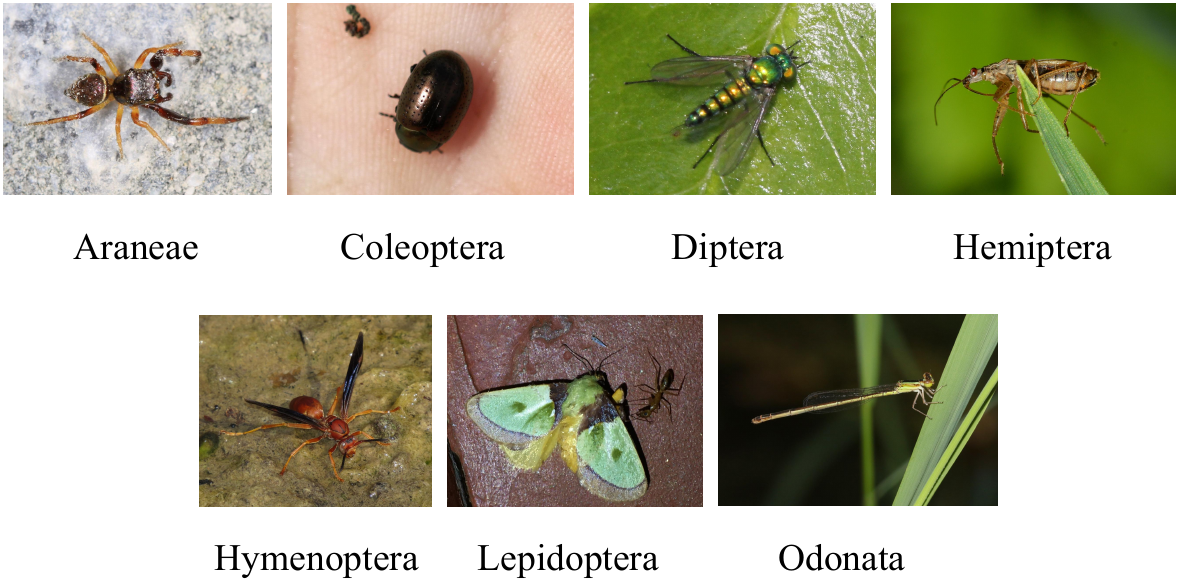}
    \caption{Examples of images and categories from ArTaxOr.}
    \label{fig:ArTaxOr}
\end{figure*}

As depicted in Figure~\ref{fig:ArTaxOr}, the ArTaxOr Dataset~\cite{Drange2020arthropod} comprises seven distinct categories of insects. The images within this dataset were primarily captured using a macro lens, resulting in high-definition visuals and a distinct foreground-background boundary. Nevertheless, the inter-class differences within the ArTaxOr Dataset are relatively subtle, posing a challenge in accurately distinguishing between the various insect categories~\cite{Xiong2023CDFSOD}. The detailed categories list and the corresponding rich texts are available in Table~\ref{tab:rich text ArTaxOr UODD1}.

\begin{figure*}[t]
    \centering
    \includegraphics[width=0.7\linewidth]{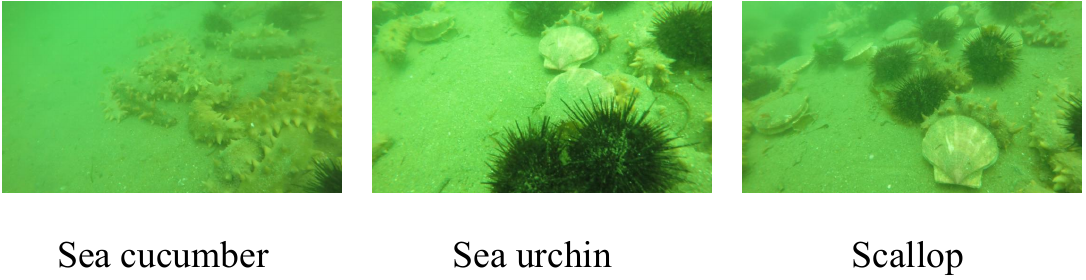}
    \caption{Examples of images and categories from UODD.}
    \label{fig:UODD}
\end{figure*}

\begin{figure*}[t]
    \centering
    \includegraphics[width=0.9\linewidth]{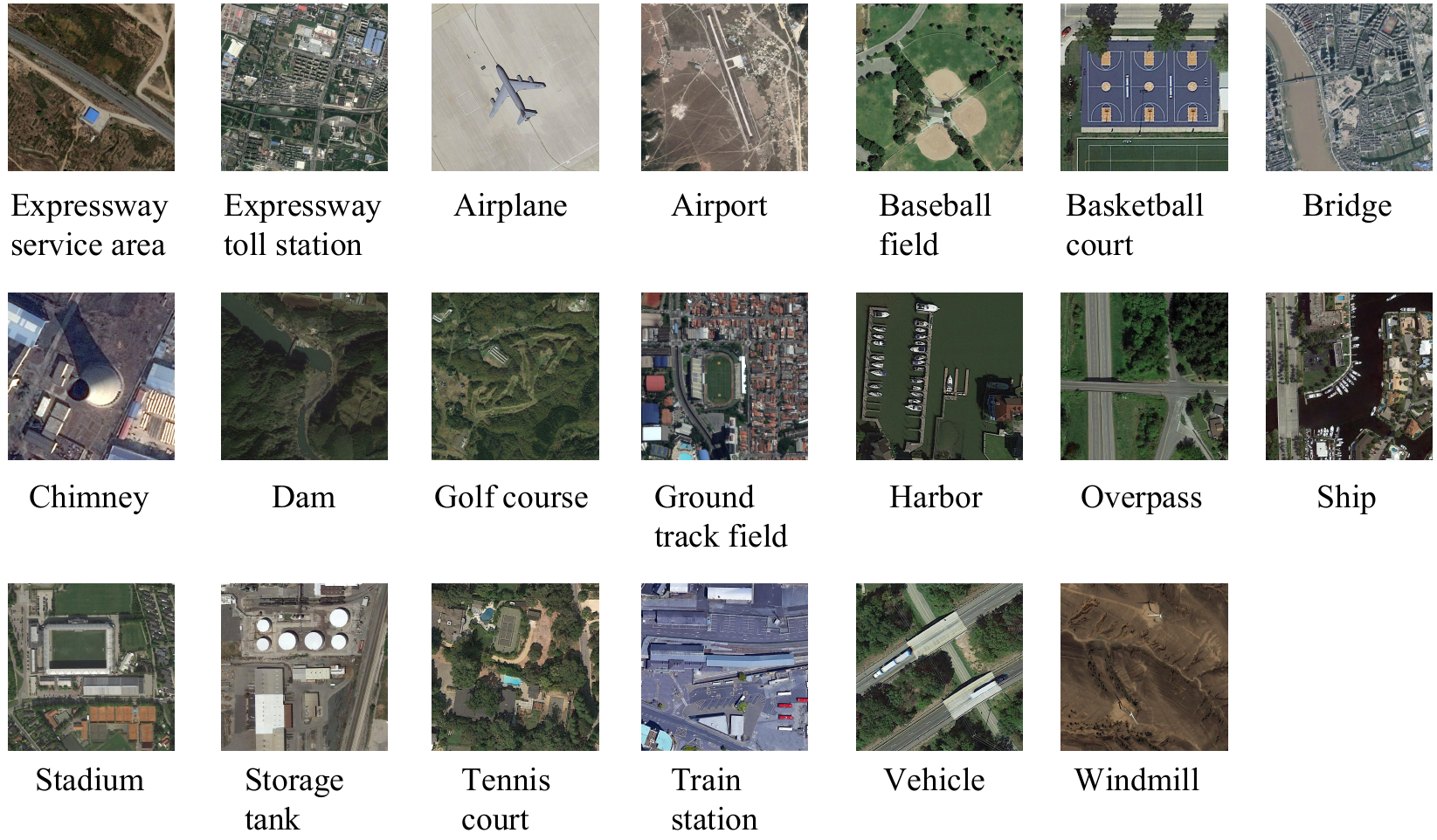}
    \caption{Examples of images and categories from DIOR.}
    \label{fig:DIOR}
\end{figure*}

\begin{figure*}[t]
    \centering
    \includegraphics[width=0.8\linewidth]{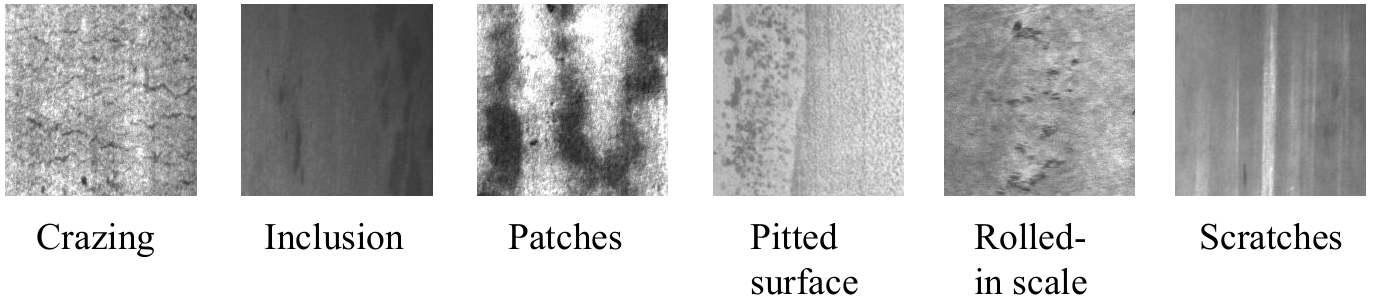}
    \caption{Examples of images and categories from NEU-DET.}
    \label{fig:NEU-DET}
\end{figure*}

\begin{figure*}[t]
    \centering
    \includegraphics[width=0.9\linewidth]{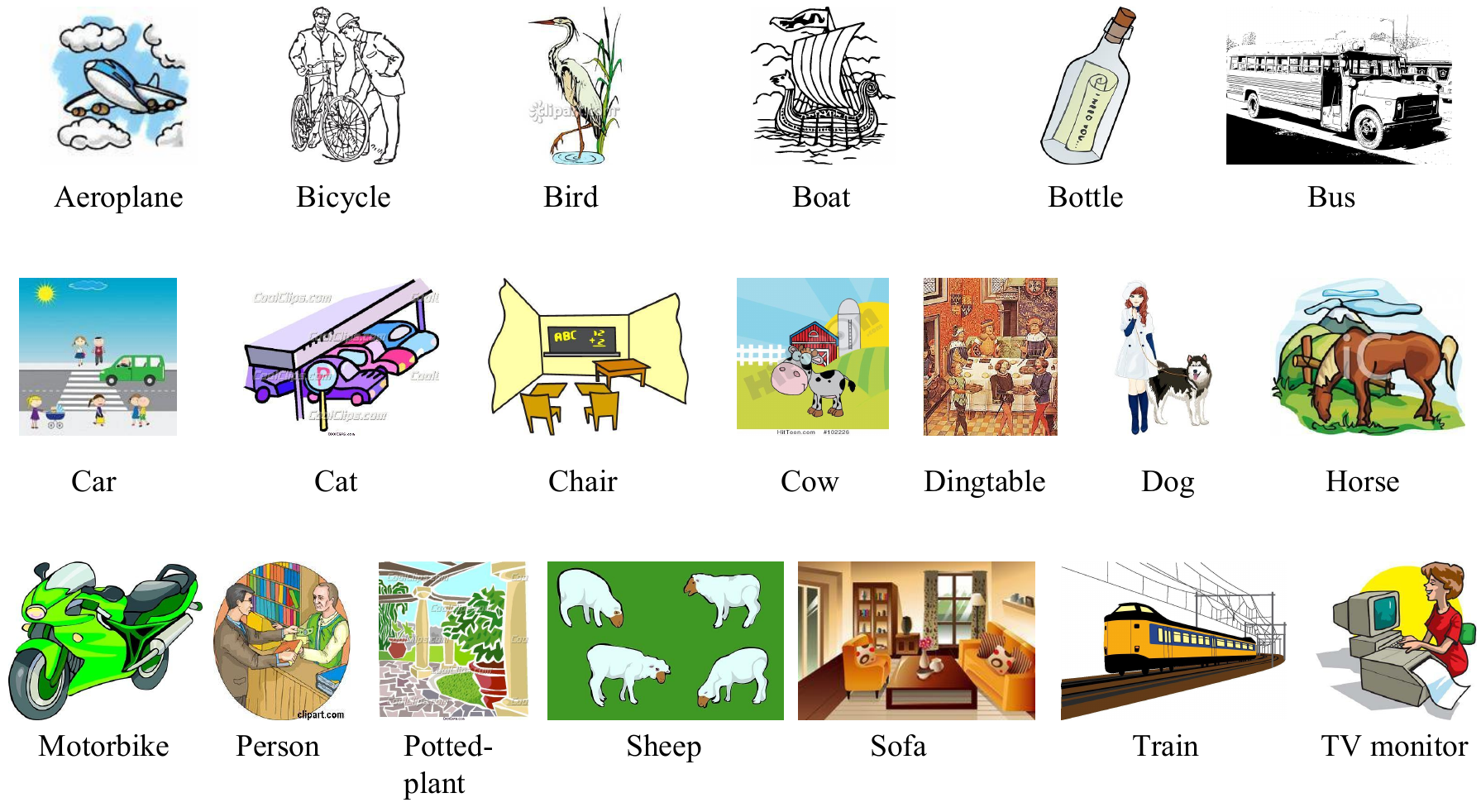}
    \caption{Examples of images and categories from Clipart1k.}
    \label{fig:Clipart1k}
\end{figure*}

\begin{figure*}[t]
    \centering
    \includegraphics[width=0.9\linewidth]{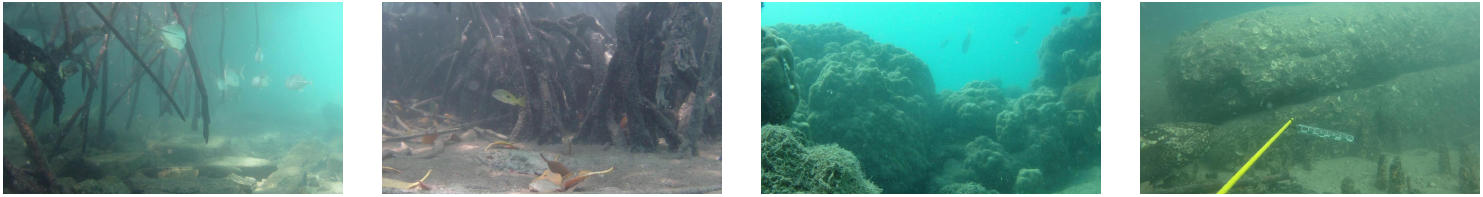}
    \caption{Examples of images and categories from DeepFish.}
    \label{fig:DeepFish}
\end{figure*}

\begin{figure*}[t]
    \centering
    \includegraphics[width=0.9\linewidth]{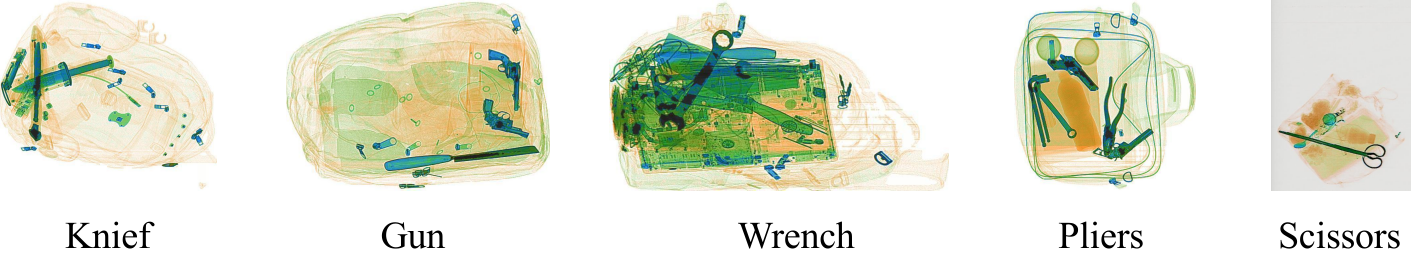}
    \caption{Examples of images and categories from SIXray.}
    \label{fig:SIXray}
\end{figure*}

\paragraph{UODD Dataset}

The UODD dataset~\cite{Jiang2021UODD}, as illustrated in Figure~\ref{fig:UODD}, comprises three distinct categories of underwater creatures. Due to the low visual contrast, it exhibits significant differences from the COCO dataset~\cite{Xiong2023CDFSOD}. The detailed categories list and the corresponding rich texts are available in Table~\ref{tab:rich text UODD2 DIOR1}.

\paragraph{DIOR Dataset}

As depicted in Figure~\ref{fig:DIOR}, the DIOR dataset~\cite{LI2020DIOR} encompasses 20 categories of Aerial scenes. These scenes differ significantly from those in the COCO dataset due to their unique remote sensing perspective, captured from a top-down shot~\cite{Xiong2023CDFSOD}. The detailed categories list and the corresponding rich texts are available in Table~\ref{tab:rich text UODD2 DIOR1},~\ref{tab:rich text DIOR2} and~\ref{tab:rich text DIOR3 and NEU-DET1}.

\paragraph{NEU-DET Dataset}

As shown in Figure~\ref{fig:NEU-DET}, the NEU-DET dataset~\cite{song2013noise} features six categories of common steel rolling product defects. These images were captured using an electron microscope, resulting in a distinct black-and-white style. Notably, the industrial image style of the NEU-DET dataset differs significantly from the life scenes presented in the COCO datasets. The detailed categories list and the corresponding rich texts are available in Table~\ref{tab:rich text DIOR3 and NEU-DET1} and~\ref{tab:rich text NEU-DET2, SIXray and Clipart1k1}.

\paragraph{Clipart1k Dataset}

The Clipart1k Dataset~\cite{inoue2018cross}, in Figure~\ref{fig:Clipart1k}, comprises 20 categories of commonly seen objects that overlap with the COCO categories, however, due to their cartoon-like style, models often struggle to accurately recognize them. The detailed categories list and the corresponding rich texts are available in Table~\ref{tab:rich text NEU-DET2, SIXray and Clipart1k1},~\ref{tab:rich text Clipart1k2},~\ref{tab:rich text Clipart1k3},~\ref{tab:rich text Clipart1k4} and~\ref{tab:rich text Clipart1k5 and DeepFish}.

\paragraph{DeepFish Dataset}

The DeepFish dataset~\cite{saleh2020realistic}, as illustrated in Figure~\ref{fig:DeepFish}, exclusively features a single category of high-definition underwater tropical fishes, all sampled in Australia. Its unique image style, characterized by blurred visual boundaries and the fishes' biological camouflage, sets it apart from the COCO dataset. The detailed categories list and the corresponding rich texts are available in Table~\ref{tab:rich text Clipart1k5 and DeepFish}.

\paragraph{SIXray Dataset}

The SIXray dataset~\cite{Miao2019SIXray}, as illustrated in Figure~\ref{fig:DeepFish}, exclusively features six category of contraband for security check. Its unique X-ray image style, characterized by complex boundaries, sets it apart from the COCO dataset.The detailed categories list and the corresponding rich texts are available in Table~\ref{tab:rich text NEU-DET2, SIXray and Clipart1k1}.

\begin{table}[t]
\centering
\begin{tabular}{l|p{1.5cm}|p{1cm}|p{9cm}}
\toprule
\textbf{Dataset} & \textbf{Category} & \multicolumn{2}{c}{\textbf{Rich text}} \\
\midrule
\multirow{21}{*}{\textbf{ArTaxOr}} & \multirow{3}{*}{Araneae} & Manually built & Araneae have eight limbs and usually perch on the silk, and do not have antennae. \\
\cmidrule(rl){3-4}
 & & LLM-aided & Araneae, commonly known as spiders, exhibit a diverse range of appearances, varying from small and delicate to large and imposing, with eight legs, two body segments, and often distinctive markings and patterns on their abdomens. \\
\cmidrule(rl){2-4}
& \multirow{4}{*}{Coleoptera} & Manually built & Beetles can usually be recognized by their two pairs of wings; the front pair is modified into horny covers that hide the rear pair and most of the abdomen and usually meet down the back in a straight line. \\
\cmidrule(rl){3-4}
& & LLM-aided & Coleoptera, the beetles, exhibit a remarkable diversity in appearance, ranging from small, shiny, and flightless to large, winged, and brightly colored, often with hardened forewings adapted for protection or flight. \\
\cmidrule(rl){2-4}
& \multirow{4}{*}{Diptera} & Manually built & Diptera use only a single pair of wings to fly, and have a mobile head, with a pair of large compound eyes, and mouthparts designed for piercing and sucking. \\
\cmidrule(rl){3-4}
& & LLM-aided & Diptera, or flies, are small insects with two wings, large compound eyes, and long antennae, varying in color from dull to bright, with some species having patterns or stripes on their bodies. \\
\cmidrule(rl){2-4}
& \multirow{3}{*}{Hemiptera} & Manually built & Hymenoptera usually have two pairs of membranous wings and the forewings are larger than the hind wings. \\
\cmidrule(rl){3-4}
& & LLM-aided & Hymenoptera are distinguished by their thin waist connecting the thorax and abdomen, as well as their typically transparent and membranous wings. \\
\cmidrule(rl){2-4}
& \multirow{3}{*}{Lepidoptera} & Manually built & Lepidoptera have scales that cover the bodies, large triangular wings, and a proboscis for siphoning nectars. \\
\cmidrule(rl){3-4}
& & LLM-aided & Lepidoptera, including butterflies and moths, are characterized by their wings, which are covered in scales and often display patterns and colors ranging from subtle to vivid. Their bodies are typically slender, and they have antennae and long, thin legs. \\
\cmidrule(rl){2-4}
& \multirow{4}{*}{Odonata} & Manually built & Odonata characteristically have large rounded heads covered mostly by compound eyes, two pairs of long, transparent wings that move independently, elongated abdomens, three ocelli and short antennae. \\
\cmidrule(rl){3-4}
& & LLM-aided & Odonata, commonly known as dragonflies and damselflies, are characterized by their large, often brightly colored eyes, long, thin bodies, and two pairs of wings, the hindwings being broader than the forewings. \\

\midrule
\multirow{10}{*}{\textbf{UODD}} & \multirow{5}{*}{Sea cucumber} & Manually built & Sea cucumbers have sausage-shape, usually resemble caterpillars; their mouth is surrounded by tentacles. \\
\cmidrule(rl){3-4}
& & Extended & Sea cucumbers have sausage-shape, usually resemble caterpillars; their mouth is surrounded by tentacles; usually seen together with sea urchins. \\
\cmidrule(rl){3-4}
& & LLM-aided & Sea cucumbers are marine invertebrates known for their elongated, leathery bodies, which are typically covered in spines or tentacles and lack a distinct head or tail. \\
\cmidrule(rl){2-4}
& \multirow{4}{*}{Sea urchin} & Manually built & Sea urchins have globular body and a radial arrangement of organs. \\
\cmidrule(rl){3-4}
& & Extended & Sea urchins have globular body and a radial arrangement of organs; usually seen together with sea cucumbers. \\
\bottomrule
\end{tabular}
\caption{Rich text of the ArTaxOr and UODD datasets}
\label{tab:rich text ArTaxOr UODD1}
\end{table}
\begin{table}[t]
\centering
\begin{tabular}{l|p{1.5cm}|p{1cm}|p{9cm}}
\toprule
\textbf{Dataset} & \textbf{Category} & \multicolumn{2}{c}{\textbf{Rich text}} \\
\midrule
\multirow{8}{*}{\textbf{UODD}} & & LLM-aided & Sea urchins are marine invertebrates characterized by their globular shape, covered in dense clusters of spines for protection, and having a distinctive oral surface with five radiating rows of teeth-like structures called pedicellariae. \\
\cmidrule(rl){2-4}
& \multirow{5}{*}{Scallop} & Manually built & The shell of a scallop has the classic fanned-out shape; one side of the shell is slightly flatter, and the other more concave in shape. \\
\cmidrule(rl){3-4}
& & Extended & Sea cucumbers have sausage-shape, usually resemble caterpillars; their mouth is surrounded by tentacles; usually seen together with sea urchins. \\
\cmidrule(rl){3-4}
& & LLM-aided & Scallops are marine bivalve mollusks characterized by their flattened, oval-shaped shells, typically with distinctive radial ribbing and a wavy or scalloped edge. \\
\midrule
\multirow{30}{*}{\textbf{DIOR}} & \multirow{4}{*}{\makecell[l]{Expressway \\ service area}} & Manually built & A expressway service area is an open space located alongside the highway, typically featuring a main building and multiple regularly arranged parking slots. \\
\cmidrule(rl){3-4}
 & & LLM-aided & Expressway service areas appear as clusters of buildings and facilities along the highway, typically surrounded by parking lots and green spaces. \\
\cmidrule(rl){2-4}
& \multirow{3}{*}{\makecell[l]{Expressway \\ toll station}} & Manually built & Toll stations on expressways are elongate, thin buildings that span across the highway, and allow vehicles to queue up for entry or exit. \\
\cmidrule(rl){3-4}
& & LLM-aided & Expressway toll stations appear as structures located along the highway, typically with multiple lanes entry and exit points, and are distinguished by their regular layout, clear lanes markings, and sometimes by the presence of toll plazas or gantries. \\
\cmidrule(rl){2-4}
& \multirow{4}{*}{Airplane} & Manually built & An airplane is equipped with a pair of symmetric, large wings positioned at the center of its cylindrical fuselage, along with a pair of smaller wings referred to as the horizontal stabilizer and a vertical stabilizer located at the tail end. \\
\cmidrule(rl){3-4}
& & LLM-aided & Airplanes are winged vehicles with a fuselage, wings, and engines, typically having a pointed nose and tail. \\
\cmidrule(rl){2-4}
& \multirow{4}{*}{Airports} & Manually built & Airports are designated open areas where airplanes can park, and they typically feature multiple comb-shaped boarding bridges, tall control towers, and lawns, but lack trees. \\
\cmidrule(rl){3-4}
& & LLM-aided & Airports are large open spaces with multiple runways, terminals, and parking lots, typically surrounded by fences and gates, and equipped with aircrafts, control towers. \\
\cmidrule(rl){2-4}
& \multirow{4}{*}{\makecell[l]{Baseball \\ field}} & Manually built & The baseball field features a diamond-shaped infield, resembling a quarter of a circle, covered with a checkerboard pattern of natural green grass. \\
\cmidrule(rl){3-4}
& & LLM-aided & A baseball field is characterized by a diamond-shaped infield with four bases, a pitcher's mound, and outfield fences, surrounded by stands for spectators and often marked by lines indicating playing areas and distances. \\
\cmidrule(rl){2-4}
& \multirow{4}{*}{\makecell[l]{Basketball \\ court}} & Manually built & The basketball court is the playing surface, consisting of a rectangular floor with tiles at either end, usually made out of a wood, often maple, and highly polished. \\
\cmidrule(rl){3-4}
& & LLM-aided & A basketball court is a rectangular playing surface with a raised rim and net at each end, typically surrounded by lines dividing the court into various zones and marked by a center circle and free-throw lines. \\
\bottomrule
\end{tabular}
\caption{Rich text of the UODD and DIOR dataset.}
\label{tab:rich text UODD2 DIOR1}
\end{table}

\begin{table}[t]
\centering
\begin{tabular}{l|p{1.5cm}|p{1cm}|p{9cm}}
\toprule
\textbf{Dataset} & \textbf{Category} & \multicolumn{2}{c}{\textbf{Rich text}} \\
\midrule
\multirow{45}{*}{\textbf{DIOR}} & \multirow{3}{*}{Bridge} & Manually built & Bridges are elongate, slender structures spanning across water bodies and may accommodate the passage of vehicles. \\
\cmidrule(rl){3-4}
& & LLM-aided & Bridges are structures built to span rivers, valleys, or other obstacles, typically with one or more arches, trusses, or cables, connecting two or more points of land. \\
\cmidrule(rl){2-4}
& \multirow{3}{*}{Chimney} & Manually built & A chimney is like a pipe or a tunnel-like channel made by metal or concrete, typically tall and expel gas. \\
\cmidrule(rl){3-4}
& & LLM-aided & Chimneys are vertical structures designed to vent smoke and gases from heating appliances, typically made of masonry or metal and having a rectangular or circular cross-section. \\
\cmidrule(rl){2-4}
& \multirow{4}{*}{Dam} & Manually built & A dam is a structure constructed across a stream, river, or estuary to retain water, and from an overhead perspective, it typically appears as a thin, curved white line. \\
\cmidrule(rl){3-4}
& & LLM-aided & Dams are large structures built across rivers or streams to control water flow, typically with a wall or gates to hold back water and create a reservoir, and may include spillways to control flooding. \\
\cmidrule(rl){2-4}
& \multirow{4}{*}{\makecell[l]{Golf \\ course}} & Manually built & Golf courses are extensive, irregularly shaped open spaces characterized by meticulously manicured lawns, and the terrain is typically gently undulating. \\
\cmidrule(rl){3-4}
& & LLM-aided & Golf courses appear as large, open spaces with regularly spaced tee boxes, greens, and fairways, typically surrounded by fences or hedges and distinguished by their smooth, manicured surfaces. \\
\cmidrule(rl){2-4}
& \multirow{4}{*}{\makecell[l]{Ground \\ track field}} & Manually built & Ground track fields typically consist of red rings forming elliptic paths, with a pair of straight lines facing each other and another pair of half-circles extending in the opposite direction. \\
\cmidrule(rl){3-4}
 & & LLM-aided & Ground track fields appear as large, open spaces with straight lines and rectangles marking the field boundaries and flight paths, typically surrounded by fences or other markers and distinguished by their regularity and symmetry. \\
\cmidrule(rl){2-4}
& \multirow{4}{*}{Harbor} & Manually built & A harbor refers to a sheltered body of water, either natural or man-made, and it is commonly seen filled with boats or ships parked within its confines. \\
\cmidrule(rl){3-4}
& & LLM-aided & Harbors appear as clusters of ships and boats moored along the coast or riverbanks, typically surrounded by docks, piers, and other facilities for loading and unloading cargo, and distinguished by their dense concentration of vessels and associated infrastructure. \\
\cmidrule(rl){2-4}
& \multirow{4}{*}{Overpass} & Manually built & Overpasses are intricate city roadways that resemble bridges and intersect each other, typically elevated above standard road levels to facilitate the passage of vehicles. \\
\cmidrule(rl){3-4}
& & LLM-aided & Overpasses appear as elevated structures spanning roads or railroads, typically with multiple arches or girders supporting the roadway above the traffic below, and distinguished by their raised position and distinctive shape. \\
\cmidrule(rl){2-4}
& \multirow{4}{*}{Ship} & Manually built & Ships typically feature a long, thin, bullet-shaped hull that floats on the water, with a sharp prow and a flat stern. \\
\cmidrule(rl){3-4}
& & LLM-aided & Ships appear as large, elongated vessels floating on the water surface, typically with a distinctive silhouette and various features such as decks, masts. \\
\bottomrule
\end{tabular}
\caption{Rich text of the DIOR dataset.}
\label{tab:rich text DIOR2}
\end{table}
\begin{table}[t]
\centering
\begin{tabular}{l|p{1.5cm}|p{1cm}|p{9cm}}
\toprule
\textbf{Dataset} & \textbf{Category} & \multicolumn{2}{c}{\textbf{Rich text}} \\
\midrule
\multirow{30}{*}{\textbf{DIOR}} & \multirow{4}{*}{Stadium} & Manually built & Stadiums resemble bowl-shaped buildings and typically feature stands surrounding a central field. \\
\cmidrule(rl){3-4}
& & LLM-aided & Stadiums appear as large, open spaces with distinctive curved or rectangular shapes, typically surrounded by stands and other facilities for spectators. \\
\cmidrule(rl){2-4}
& \multirow{4}{*}{\makecell[l]{Storage \\ tank}} & Manually built & Storage tanks are usually large cylindrical or spherical metal containers with a clean exterior, clustered and well-arranged on the ground. \\
\cmidrule(rl){3-4}
& & LLM-aided & Storage tanks appear as large, circular or rectangular structures with flat roofs, typically surrounded by fenced areas and distinguished by their regular shape, smooth surfaces, and sometimes by the presence of multiple tanks clustered together. \\
\cmidrule(rl){2-4}
& \multirow{3}{*}{\makecell[l]{Tennis \\ court}} & Manually built & Tennis courts are rectangular fields marked with straight lines across their surfaces, and they are typically colored green or blue. \\
\cmidrule(rl){3-4}
& & LLM-aided & Tennis courts appear as rectangular areas with distinctive lines marking the boundaries and service lines, typically surrounded by fences or hedges and distinguished by their regularity and symmetry. \\
\cmidrule(rl){2-4}
& \multirow{4}{*}{\makecell[l]{Train \\ station}} & Manually built & Train stations are large buildings designed to accommodate trains, and they typically feature a dense network of multiple railways passing through them. \\
\cmidrule(rl){3-4}
& & LLM-aided & Train stations appear as clusters of buildings and platforms along the railway tracks, typically with multiple tracks converging and diverging, and distinguished by the presence of trains or other rail vehicles. \\
\cmidrule(rl){2-4}
& \multirow{3}{*}{Vehicle} & Manually built & Vehicles are typically small rectangular objects commonly seen traveling along roads or parked in parking lots. \\
\cmidrule(rl){3-4}
& & LLM-aided & Vehicles appear as small to medium-sized objects with distinctive shapes and features, such as wheels, windows, and roofs, typically moving along roads or parking areas. \\
\cmidrule(rl){2-4}
& \multirow{3}{*}{Windmill} & Manually built & Windmills are tall structures topped with multiple fan-like blades that spread out and rotate to harness wind energy. \\
\cmidrule(rl){3-4}
& & LLM-aided & Windmills appear as tall, thin structures with three or more long blades rotating in the wind, typically located in open spaces such as fields or coastal areas. \\
\midrule
\multirow{12}{*}{\textbf{NEU-DET}} & \multirow{2}{*}{\makecell[l]{Rolled-in \\ scale}} & Manually built & Rolled-in scale typically consists of multiple rough patches arranged in a fish scale-like pattern. \\
\cmidrule(rl){3-4}
 & & LLM-aided & In hot-rolled steel strip defect images, rolled-in scale appears as dark, irregular patches or streaks along the surface of the steel strip. \\
\cmidrule(rl){2-4}
& \multirow{2}{*}{Patches} & Manually built & Patch defects typically appear as dark, dense clusters. \\
\cmidrule(rl){3-4}
& & LLM-aided & In hot-rolled steel strip defect images, a patch typically appears as a localized area of abnormal texture, color, or brightness. \\
\cmidrule(rl){2-4}
& \multirow{3}{*}{Crazing} & Manually built & Crazings typically consist of numerous tiny, thin lines resembling tree branches. \\
\cmidrule(rl){3-4}
& & LLM-aided & Crazing in hot-rolled steel strip defect images is characterized by a network of fine cracks or fissures that appear on the surface of the steel strip, typically have a regular, hairline pattern. \\
\bottomrule
\end{tabular}
\caption{Rich text of the DIOR and NEU-DET dataset.}
\label{tab:rich text DIOR3 and NEU-DET1}
\end{table}
\begin{table}[t]
\centering
\begin{tabular}{l|p{1cm}|p{1cm}|p{9cm}}
\toprule
\textbf{Dataset} & \textbf{Category} & \multicolumn{2}{c}{\textbf{Rich text}} \\
\midrule
\multirow{12}{*}{\textbf{NEU-DET}} & \multirow{2}{*}{\makecell[l]{Pitted \\ surface}} & Manually built & Pitted surfaces are characterized by a dense cluster of tiny spots. \\
\cmidrule(rl){3-4}
& & LLM-aided & Pitted surfaces are characterized by numerous small, circular indentations or depressions spread across the steel strip's surface. \\
\cmidrule(rl){2-4}
& \multirow{5}{*}{Inclusion} & Manually built & Inclusions typically consist of a series of dark, slender patches. \\
\cmidrule(rl){3-4}
& & LLM-aided & Inclusions in hot-rolled steel strip defect images appear as dark, irregularly shaped areas within the steel strip's matrix, typically caused by foreign particles or impurities during the manufacturing. \\
\cmidrule(rl){2-4}
& \multirow{2}{*}{Scratches} & Manually built & Scratches are typically characterized by multiple light, straight, slender lines. \\
\cmidrule(rl){3-4}
& & LLM-aided & Scratches appear as long, continuous lines or marks across the steel strip's surface. \\

\midrule
\multirow{14}{*}{\textbf{SIXray}} & \multirow{2}{*}{Knife} & Manually built & Knives appear as long, dark shadows, consisting of a handle and a sharp blade. \\
\cmidrule(rl){3-4}
 & & LLM-aided & The knife typically exhibits a blade with a sharp edge for cutting. \\
\cmidrule(rl){2-4}
& \multirow{2}{*}{Gun} & Manually built & Guns are discernible as dark L-shaped shadows, featuring a long, slender barrel on one side and a shorter, stubby handle on the other. \\
\cmidrule(rl){3-4}
& & LLM-aided & The gun typically exhibits a compact and handgun-like shape, featuring a barrel, a grip for holding, a trigger for firing. \\
\cmidrule(rl){2-4}
& \multirow{2}{*}{Wrench} & Manually built & The wrench is discernible as a slim, elongated shape, with one end featuring a horseshoe-shaped head and the other end attached with a circular ring. \\
\cmidrule(rl){3-4}
& & LLM-aided & The wrench typically features a handle for gripping and a jaw or socket that can be adjusted to fit nuts, bolts, or other fasteners, enabling it to rotate or tighten them. \\
\cmidrule(rl){2-4}
& \multirow{2}{*}{Pliers} & Manually built & Pliers are characterized by their two slender handles, connected to a pointed, snip-nosed jaw. \\
\cmidrule(rl){3-4}
& & LLM-aided & The pliers typically consist of two pivoting levers with jaws at the ends, designed to grasp and manipulate objects, with handles for easy gripping and control. \\
\cmidrule(rl){2-4}
& \multirow{2}{*}{Scissors} & Manually built & Scissors comprise two circular handles connected to a long, sharp blade that forms the cutting jaw. \\
\cmidrule(rl){3-4}
& & LLM-aided & The scissors typically consist of two pivoting blades with sharp edges that meet at a point, allowing for cutting action when the handles are squeezed together. \\
\midrule
\multirow{6}{*}{\textbf{Clipart1k}} & \multirow{4}{*}{Aeroplane} & Manually built & The cartoon aeroplane typically features a pair of wings positioned in the center of its cylindrical body, accompanied by a set of small stabilizers at the rear, usually in unified colors and clear contours. \\
\cmidrule(rl){3-4}
 & & LLM-aided & A cartoon aeroplane is characterized by its sleek fuselage, wings and various mechanical components, all exaggerated in features, enhanced with brighter colors, and adorned in a whimsical or anthropomorphic design. \\
\bottomrule
\end{tabular}
\caption{Rich text of the NEU-DET, SIXray and Clipart1k dataset.}
\label{tab:rich text NEU-DET2, SIXray and Clipart1k1}
\end{table}

\begin{table}[t]
\centering
\begin{tabular}{l|p{1cm}|p{1cm}|p{9cm}}
\toprule
\textbf{Dataset} & \textbf{Category} & \multicolumn{2}{c}{\textbf{Rich text}} \\
\midrule
\multirow{39}{*}{\textbf{Clipart1k}} & \multirow{4}{*}{Bicycle} & Manually built & The cartoon bicycle typically comprises of two wheels, one at the front and another at the rear, along with a triangular-shaped frame, usually in unified colors and clear contours. \\
\cmidrule(rl){3-4}
& & LLM-aided & A cartoon bicycle is typically defined by its two wheels, a frame connecting them, handlebars for steering, and a seat for the rider, all exaggerated in features, enhanced with brighter colors, and adorned in a whimsical or anthropomorphic design. \\
\cmidrule(rl){2-4}
& \multirow{4}{*}{Bird} & Manually built & The cartoon boat typically features a half-moon-shaped hull floating on the water, and sometimes with a set of big sails, usually in unified colors and clear contours. \\
\cmidrule(rl){3-4}
& & LLM-aided & A cartoon bird is typically characterized by its feathered body, beaked head, and wings, all exaggerated in features, enhanced with brighter colors, and adorned in a whimsical or anthropomorphic design. \\
\cmidrule(rl){2-4}
& \multirow{4}{*}{Boat} & Manually built & The cartoon bottle commonly exhibits a transparent cylindrical body, characterized by a slender neck and a wider abdomen, usually in unified colors and clear contours. \\
\cmidrule(rl){3-4}
& & LLM-aided & A cartoon boat is typically defined by its hull, which floats on water, along with features such as a deck, mast, and sails or engines, all exaggerated in features, enhanced with brighter colors, and adorned in a whimsical or anthropomorphic design. \\
\cmidrule(rl){2-4}
& \multirow{4}{*}{Bottle} & Manually built & The cartoon bottle commonly exhibits a transparent cylindrical body, characterized by a slender neck and a wider abdomen, usually in unified colors and clear contours. \\
\cmidrule(rl){3-4}
& & LLM-aided & A cartoon bottle is typically characterized by its distinct shape, often cylindrical or tapered, with a narrow neck and a mouth for pouring or sealing, all exaggerated in features, enhanced with brighter colors, and adorned in a whimsical or anthropomorphic design. \\
\cmidrule(rl){2-4}
& \multirow{5}{*}{Bus} & Manually built & The cartoon bus typically takes the form of a cuboid-shaped vehicle, supported by four wheels at the base. Its body is surrounded by rectangular windows, usually in unified colors and clear contours. \\
\cmidrule(rl){3-4}
& & LLM-aided & A cartoon bus is typically characterized by its large size, rectangular shape, and multiple doors for passengers, along with windows for visibility and a roof that often features an advertising panel or other decorative elements, all exaggerated in features, enhanced with brighter colors, and adorned in a whimsical or anthropomorphic design. \\
\cmidrule(rl){2-4}
& \multirow{5}{*}{Car} & Manually built & Cartoon cars are vehicles typically featuring a small, box-shaped driving cab and an flat elongated front engine box, supported by four wheels at the base, usually in unified colors and clear contours. \\
\cmidrule(rl){3-4}
& & LLM-aided & A cartoon car is typically defined by its sleek and compact body, with four wheels, a windshield, and various features such as headlights, taillights, and mirrors, all exaggerated in features, enhanced with brighter colors, and adorned in a whimsical or anthropomorphic design. \\
\bottomrule
\end{tabular}
\caption{Rich text of the Clipart1k dataset.}
\label{tab:rich text Clipart1k2}
\end{table}

\begin{table}[t]
\centering
\begin{tabular}{l|p{1.5cm}|p{1cm}|p{8cm}}
\toprule
\textbf{Dataset} & \textbf{Category} & \multicolumn{2}{c}{\textbf{Rich text}} \\
\midrule
\multirow{39}{*}{\textbf{Clipart1k}} & \multirow{4}{*}{Cat} & Manually built & Cartoon cats possess a furry body with four limbs, a pair of triangular ears on top of their heads, long tails, and slightly pointed mouths, usually in unified colors and clear contours. \\
\cmidrule(rl){3-4}
& & LLM-aided & A cartoon cat is typically characterized by its sleek and muscular body, furry coat, pointed ears, and large, expressive eyes, all exaggerated in features, enhanced with brighter colors, and adorned in a whimsical or anthropomorphic design. \\
\cmidrule(rl){2-4}
& \multirow{4}{*}{Chair} & Manually built & The cartoon chair typically features four slender legs for support, along with a small, horizontal, flat area designed for sitting. sometimes includes a vertical backrest and a pair of armrests, usually in unified colors and clear contours. \\
\cmidrule(rl){3-4}
& & LLM-aided & A cartoon chair is typically defined by its supportive structure, consisting of a seat, backrest, and legs, all exaggerated in features, enhanced with brighter colors, and adorned in a whimsical or anthropomorphic design. \\
\cmidrule(rl){2-4}
& \multirow{4}{*}{Cow} & Manually built & The cartoon cow commonly possesses four long legs that support its box-shaped body, with a long tail, sometimes, it may also feature a pair of crescent-shaped horns protruding from its head, usually in unified colors and clear contours. \\
\cmidrule(rl){3-4}
& & LLM-aided & A cartoon cow is typically characterized by its large, stocky body, brown or black fur, and prominent horns, all exaggerated in features, enhanced with brighter colors, and adorned in a whimsical or anthropomorphic design. \\
\cmidrule(rl){2-4}
& \multirow{4}{*}{Diningtable} & Manually built & The cartoon dining table is typically a large, tall piece of furniture supported by multiple legs; it's surface is often rounded or rectangular, usually in unified colors and clear contours. \\
\cmidrule(rl){3-4}
& & LLM-aided & A cartoon dining table is typically defined by its rectangular or circular shape, flat surface for placing food and dishes, and supportive legs, all exaggerated in features, enhanced with brighter colors, and adorned in a whimsical or anthropomorphic design. \\
\cmidrule(rl){2-4}
& \multirow{4}{*}{Dog} & Manually built & The cartoon dog typically exhibits a furry body, characterized by a prominent, elongated nose and mouth, leaf-shaped ears, four legs, and a tail that is often curled, usually in unified colors and clear contours. \\
\cmidrule(rl){3-4}
& & LLM-aided & A cartoon dog is typically characterized by its varied coats, distinct ears and muzzle shapes, as well as its four legs and tail, all exaggerated in features, enhanced with brighter colors, and adorned in a whimsical or anthropomorphic design. \\
\cmidrule(rl){2-4}
& \multirow{4}{*}{Horse} & Manually built & The cartoon horse typically stands on four legs, supporting its body, and features a brush-like tail, a long neck and a long face, as well as a long mane along its neck, usually in unified colors and clear contours. \\
\cmidrule(rl){3-4}
& & LLM-aided & A cartoon horse is typically characterized by its powerful legs, long mane and tail, and muscular body, all exaggerated in features, enhanced with brighter colors, and adorned in a whimsical or anthropomorphic design. \\
\bottomrule
\end{tabular}
\caption{Rich text of the Clipart1k dataset.}
\label{tab:rich text Clipart1k3}
\end{table}

\begin{table}[t]
\centering
\begin{tabular}{l|p{1.5cm}|p{1cm}|p{8cm}}
\toprule
\textbf{Dataset} & \textbf{Category} & \multicolumn{2}{c}{\textbf{Rich text}} \\
\midrule
\multirow{42}{*}{\textbf{Clipart1k}} & \multirow{4}{*}{Motorbike} & Manually built & The cartoon motorbike typically features two bold wheels at the front and rear of its body, and a prominent engine positioned in the middle, usually in unified colors and clear contours. \\
\cmidrule(rl){3-4}
& & LLM-aided & A cartoon motorbike is typically defined by its compact and aerodynamic frame, two wheels, a handlebar for steering, and an engine for power, all exaggerated in features, enhanced with brighter colors, and adorned in a whimsical or anthropomorphic design. \\
\cmidrule(rl){2-4}
& \multirow{4}{*}{Person} & Manually built & The cartoon person typically possesses two legs to support their body, a pair of arms, and a head with a pair of eyes, ears, and a mouth, usually in unified colors and clear contours. \\
\cmidrule(rl){3-4}
& & LLM-aided & A cartoon person is typically characterized by a bipedal stance, upright posture, and a head with facial features such as eyes, nose, and mouth, along with arms, legs, and varying skin tones and hairstyles, all exaggerated in features, enhanced with brighter colors, and adorned in a whimsical or anthropomorphic design. \\
\cmidrule(rl){2-4}
& \multirow{4}{*}{Potted plant} & Manually built & The cartoon potted plants are typically small trees or bushes grown indoors in pots, with green leaves or vibrant flowers, usually in unified colors and clear contours. \\
\cmidrule(rl){3-4}
& & LLM-aided & A cartoon potted plant is typically defined by its green foliage or flowers, growing from a soil-filled pot, all exaggerated in features, enhanced with brighter colors, and adorned in a whimsical or anthropomorphic design. \\
\cmidrule(rl){2-4}
& \multirow{4}{*}{Sheep} & Manually built & The cartoon sheep commonly stands on four legs, supporting its body, and is covered in white, curly fur; it features a short, curly tail and a pair of horns that curl downward, usually in unified colors and clear contours. \\
\cmidrule(rl){3-4}
& & LLM-aided & A cartoon sheep is typically characterized by its fleecy white coat, rounded body, and curved horns, all exaggerated in features, enhanced with brighter colors, and adorned in a whimsical or anthropomorphic design. \\
\cmidrule(rl){2-4}
& \multirow{4}{*}{Sofa} & Manually built & The cartoon sofa boasts a long, box-shaped body designed for seating, adorned with cushions and featuring a thick backrest as well as a pair of sturdy armrests, usually in unified colors and clear contours. \\
\cmidrule(rl){3-4}
& & LLM-aided & A cartoon sofa is typically defined by its cushioned seating, backrest, and arms, all exaggerated in features, enhanced with brighter colors, and adorned in a whimsical or anthropomorphic design. \\
\cmidrule(rl){2-4}
& \multirow{4}{*}{Train} & Manually built & The cartoon train comprises a succession of long, box-shaped compartments that traverse along a rail, usually in unified colors and clear contours. \\
\cmidrule(rl){3-4}
& & LLM-aided & A cartoon train is typically characterized by its long and connected series of carriages, each with windows and doors, a powerful locomotive at the front, all exaggerated in features, enhanced with brighter colors, and adorned in a whimsical or anthropomorphic design. \\
\bottomrule
\end{tabular}
\caption{Rich text of the Clipart1k dataset.}
\label{tab:rich text Clipart1k4}
\end{table}

\begin{table}[t]
\centering
\begin{tabular}{l|p{1.5cm}|p{1cm}|p{8cm}}
\toprule
\textbf{Dataset} & \textbf{Category} & \multicolumn{2}{c}{\textbf{Rich text}} \\
\midrule
\cmidrule(rl){2-4}
\multirow{8}{*}{\textbf{Clipart1k}} & \multirow{4}{*}{\makecell[l]{TV \\ monitor}} & Manually built & The cartoon TV monitor typically comprises a screen for displaying images and a sturdy base for support, usually in unified colors and clear contours. \\
\cmidrule(rl){3-4}
& & LLM-aided & A cartoon TV monitor is typically defined by its rectangular screen, displaying images and sounds, with a frame or border, and various buttons or controls for operation, all exaggerated in features, enhanced with brighter colors, and adorned in a whimsical or anthropomorphic design. \\

\midrule
\multirow{4}{*}{\textbf{DeepFish}} & \multirow{4}{*}{Fish} & Manually built & Underwater fish typically exhibit a spindle-shaped body covered with scales, along with a pair of pectoral fins and a tail that is crescent-shaped. \\
\cmidrule(rl){3-4}
& & LLM-aided & Underwater fish typically exhibit diverse appearance features, ranging from sleek and streamlined bodies to brightly colored scales and fins. \\
\bottomrule
\end{tabular}
\caption{Rich text of the Clipart1k and DeepFish dataset.}
\label{tab:rich text Clipart1k5 and DeepFish}
\end{table}

\subsubsection{Implementation Details}
\label{sec:Implementation}

Our experiments were conducted using hardware equipped with 8 NVIDIA 3090 GPUs to enable parallelized training and testing. We adopted hyperparameter configurations inspired by DETR~\cite{Carion20DETR}, setting the initial learning rate to 0.001. The batch size was adjusted depending on the scenario: 4 for 1-shot, and 1 for both 5-shot and 10-shot cases. The DETR encoder and decoder each consist of six self-attention layers, with the first encoder layer integrated into our meta-learning multimodal aggregation module. 
We employed DETR-R101 as the vision-language backbone, and the image-text transformer decoder architecture follows the design by Afham et al.~\cite{afham2021rich}. For transparency and reproducibility, our code is included in the supplementary materials to allow others to verify and build upon our results.

\subsubsection{Baseline For Comparison and Metrics}

In our comparative analysis, we evaluated our method against recent state-of-the-art approaches that demonstrate competitive performance in the field. To ensure a fair and consistent comparison, we selected methods widely recognized in the literature for their effectiveness. The evaluation metric used for benchmarking is the mean Average Precision (mAP), which is a standard metric for assessing object detection performance. This allowed us to quantify and highlight the improvements our approach offers relative to existing methods.

\subsection{Comparative Results}

We demonstrate that our proposed algorithm outperforms existing methods on all the four benchmarks we use in our experiments.

\subsubsection{Results for CD-FSOD}

Our experimental results on the CD-FSOD datasets, summarized in Table~\ref{tab:main}, align with the setup described by Xiong et al.~\cite{Xiong2023CDFSOD}. Specifically, our model is pre-trained on COCO and fine-tuned on three out-of-domain datasets, with mean Average Precision (mAP) serving as the evaluation metric. Since no prior work addresses CDMM-FSOD, we implemented and evaluated representative FSOD and MMOD methods, including Meta-DETR, Next-Chat, GOAT, and ViLD, on the CD-FSOD benchmark using their official code. Results from Distill-CD-FSOD~\cite{Xiong2023CDFSOD} and Fu et al.~\cite{fu2025cross} are directly cited from their papers.

Our approach significantly outperforms baseline methods, particularly on the ArTaxOr dataset, achieving mAP values of 15.1, 48.7, and 61.4 for the 1-shot, 5-shot, and 10-shot scenarios, compared to the highest baseline values of 6.5, 20.9, and 23.4, respectively. This improvement is attributed to the reduced domain shift between ArTaxOr's well-defined object boundaries and the COCO pre-training dataset, in contrast to other benchmarks. 
Additionally, leveraging LLM-generated rich text improves performance across all three datasets by providing more comprehensive descriptions. However, this advantage is not universal; for datasets like NEU-DET, LLMs occasionally fail to produce distinct descriptions for abstract or highly technical category names, limiting their effectiveness.


\begin{table*}[t]
\centering
\resizebox{1\textwidth}{!}{
\begin{tabular}{c|c|c|ccc|ccc|ccc}
\toprule
\multirow{2}{*}{\diagbox[height=20pt,innerrightsep=25pt]{\textbf{Method}}{\textbf{Shot}}} & \multirow{2}{*}{\textbf{Backbone}} & \multirow{2}{*}{\textbf{Modality}} & \multicolumn{3}{c|}{\textbf{ArTaxOr}} & \multicolumn{3}{c|}{\textbf{DIOR}} & \multicolumn{3}{c}{\textbf{UODD}} \\
			& & & 1 & 5 & 10 & 1 & 5 & 10 & 1 & 5 & 10 \\
\midrule

Meta-RCNN $\circ$ ~\cite{yan2019meta} & ResNet50 & Single & 2.8 & 8.5 & 14.0 & 7.8 & 17.7 & 20.6 & 3.6 & 8.8 & 11.2\\

TFA w/cos $\circ$~\cite{Wang20TFA} & ResNet50 & Single & 3.1 & 8.8 & 14.0 & 8.0 & 18.1 & 20.5 & 4.4 & 8.7 & 11.8 \\

FSCE $\circ$~\cite{sun2021fsce} & ResNet50 & Single &  3.7 & 10.2 & 15.9 & 8.6 & 18.7 & 21.9 & 3.9 & 9.6 & 12.0 \\ 

DeFRCN $\circ$~\cite{qiao2021defrcn} & ResNet50 & Single & 3.6 & 9.9 & 15.5 & 9.3 & 18.9 & 22.9 & 4.5 & 9.9 & 12.1 \\ 

Distill-cdfsod $\circ$~\cite{Xiong2023CDFSOD} & ResNet50 & Single & 
5.1 & 12.5 & 18.1 & 10.5 & 19.1 & 26.5 & 5.9 & 12.2 & 14.5 \\

ViTDeT-FT$\dagger$ ~\cite{li2022exploring} & ViT-B/14 & Single & 5.9 & 20.9 & 23.4 & 12.9 & 23.3 & 29.4 & 4.0 & 11.1 & 15.6 \\

Detic-FT$\dagger$ ~\cite{zhou2022detecting} & ViT-L/14 & Multi & 3.2 & 8.7 & 12.0 & 4.1 & 12.1 & 15.4 & 4.2 & 10.4 & 14.4 \\
DE-ViT$\dagger$~\cite{zhang2023detect} & ViT-L/14 & Single & 0.4 & 10.1 & 9.2 & 2.7 & 7.8 & 8.4 & 1.5 & 3.1 & 3.1 \\
Meta-DETR$\dagger$~\cite{Zhang23MetaDETR} & ViT-L/14 & Single & 6.5 & 11.3 & 14.5 & 11.1 & 19.4 & 19.9 & 5.8 & 9.5 & 9.2 \\
\midrule 
Next-Chat$\ddag$~\cite{zhang2023nextchat} & ViT-L/14 & Multi & 1.1 & 10.9 & 11.2 & 10.6 & 19.2 & 19.3 & 2.1 & 3.2 & 4.0 \\ 
GOAT$\ddag$~\cite{Wang23GOAT} & ResNet50 & Multi & 5.7 & 11.1 & 21.2 & 11.3 & 20.1 & 25.2 & 3.5 & 9.5 & 16.6 \\
ViLD$\ddag$~\cite{gu2022openvocabulary} & ResNet50 & Multi & 4.4 & 10.2 & 16.1 & 10.2 & 18.8 & 23.7 & 2.8 & 5.5 & 12.3 \\
\midrule
\textbf{Our Method} w/ self-built text & DETR-R101 & Multi & \textbf{14.8} & \textbf{48.1} & \textbf{61.0} & \textbf{13.7} & \textbf{26.3} & \textbf{31.3} & \textbf{5.5} & \textbf{11.9} & \textbf{17.4} \\
\textbf{Our Method} w/ LLM text & DETR-R101 & Multi & \textbf{15.1} & \textbf{48.7} & \textbf{61.4} & \textbf{14.3} & \textbf{26.9} & \textbf{31.4} & \textbf{6.9} & \textbf{12.5} & \textbf{17.5} \\
\bottomrule
\end{tabular}
}
\caption{Performance Results (mAP) on CD-FSOD benchmarks. The $\circ$ denotes results of general FSOD methods from Distill-cdfsod~\cite{Xiong2023CDFSOD}; $\dagger$ denotes that the methods are developed or the results are reported by Fu et al.   ~\cite{fu2025cross}; $\ddag$ indicates the results of MM-FSOD   are reported by us. Highest scores are in bold font.}
\label{tab:main}
\end{table*}

\begin{table*}[t]
	\centering
	\resizebox{1\textwidth}{!}{
		\begin{tabular}{l|l|lllll|lllll|lllll}
			\toprule
			\multirow{2}{*}{\diagbox[height=20pt,innerrightsep=25pt]{Method}{Shot}} & \multirow{2}{*}{Backbone} & \multicolumn{5}{c|}{Split1} & \multicolumn{5}{c|}{Split2} & \multicolumn{5}{c}{Split3} \\
			& & 1 & 2 & 3 & 5 & 10 & 1 & 2 & 3 & 5 & 10 & 1 & 2 & 3 & 5 & 10\\
			\midrule
    TIP \cite{Li21TIP} & FRCN-R101 & 27.7 & 36.5 & 43.3 & 50.2 & 59.6 & 22.7 & 30.1 & 33.8 & 40.9 & 46.9 & 21.7 & 30.6 & 38.1 & 44.5 & 50.9 \\
			CME \cite{Li21CME} & FRCN-R101 & 41.5 & 47.5 & 50.4 & 58.2 & 60.9 & 27.2 & 30.2 & 41.4 & 42.5 & 46.8 & 34.3 & 39.6 & 45.1 & 48.3 & 51.5 \\
			DC-Net \cite{Hu21DCNet} & FRCN-R101 & 33.9 & 37.4 & 43.7 & 51.1 & 59.6 & 23.2 & 24.8 & 30.6 & 36.7 & 46.6 & 32.3 & 34.9 & 39.7 & 42.6 & 50.7 \\
           CGDP \cite{Li21CGDP} & FRCN-R2101 & 40.7 & 45.1 & 46.5 & 57.4 & 62.4 & 27.3 & 31.4 & 40.8 & 42.7 & 46.3 & 31.2 & 36.4 & 43.7 & 50.1 & 55.6 \\
        Meta Faster R-CNN \cite{Han21MetaFRCN} & FRCN-R101 & 40.2 & 30.5 & 33.3 & 42.3 & 46.9 & 26.8 & 32.0 & 39.0 & 37.7 & 37.4 & 34.0 & 32.5 & 34.4 & 42.7 & 44.3 \\
           FCT \cite{Han22FCT} & PVTv2-B2-Li & 49.9 & \textbf{57.1} & 57.9 & 63.2 & 67.1 & 27.6 & 34.5 & 43.7 & 49.2 & 51.2 & 39.5 & \textbf{54.7} & 52.3 & 57.0 & 58.7 \\
           Meta-DETR \cite{Zhang23MetaDETR} & DETR-R101 & 40.6 & 51.4 & 58.0 & 59.2 & 63.6 & 37.0 & 36.6 & 43.7 & 49.1 & 54.6 & 41.6 & 45.9 & 52.7 & 58.9 & 60.6 \\
           FM-FSOD \cite{Han2024FMFSOD} & ViT-S & 41.6 & 49.0 & 55.8 & 61.2 & 67.7 & 34.7 & 37.6 & 47.6 & 52.5 & \textbf{58.7} & 39.5 & 47.8 & 54.4 & 57.8 & 62.6 \\
			MM-FSOD \cite{han2023multimodal} & \multirow{1}{*}{ViT} & 42.5 & 41.2 & 41.6 & 48.0 & 53.4 & 30.5 & 34.0 & 39.3 & 36.8 & 37.6 & 39.9 & 37.0 & 38.2 & 42.5 & 45.6 \\
			\midrule
			\textbf{Our Method} & DETR-R101 & \textbf{45.1} & 54.9 & \textbf{62.2} & \textbf{65.1} & \textbf{69.6} & \textbf{41.7} & \textbf{42.1} & \textbf{47.7} & \textbf{54.3} & 57.3 & \textbf{49.2} & 52.2 & \textbf{56.9} & \textbf{63.1} & \textbf{64.0} \\
			\bottomrule
		\end{tabular}
  }
\caption{Performance results on PASCAL-VOC dataset w/ self-built rich text descriptions. Highest scores are bolded.}
\label{tab:voc}
\end{table*}

\subsubsection{Results for PASCAL-VOC}
Our few-shot detection results on the PASCAL VOC dataset are presented in Table~\ref{tab:voc}. Consistent with TFA~\cite{Wang20TFA} and the setup outlined in Section~\ref{sec:datasets}, we use three fixed base-novel category splits (Split 1, 2, and 3) to minimize sampling bias. For each split, predefined $n$-shot sampling ($n \in \{1,2,3,5,10\}$) is applied to the novel categories. The model is evaluated across these dataset splits.
As detailed in Section~\ref{sec:Preliminaries}, fine-tuning also involves base categories to mitigate forgetting, but with different sampling strategies: (i) restricted $n$-shot sampling for base categories, as employed by TFA~\cite{Wang20TFA}; and (ii) unrestricted sampling for base categories, as used in Meta-DETR~\cite{Zhang23MetaDETR}. Both approaches are valid for FSOD; we adopt the latter.
As shown in Table~\ref{tab:voc}, our method demonstrates superior performance compared to prior work, highlighting its effectiveness.

\subsubsection{Results for Mo-FSOD}

The results on the Mo-FSOD benchmark are shown in Table~\ref{tab:balanced cdfsod}, where our approach outperforms previous methods, particularly on the Clipart1k dataset. In the 10-shot scenario, we achieve scores of 47.7 and 48.8, significantly higher than the best prior method's score of 25.6. However, on the NEU-DET dataset, LLM-generated text descriptions do not yield better performance than manually crafted descriptions. This is due to the LLM's inability to generate sufficiently distinctive descriptions for the six defect categories, resulting in text homogeneity that limits its effectiveness.

\subsubsection{Results for CDS-FSOD}

The results on the CDS-FSOD benchmark, presented in Table~\ref{tab:unbalanced cdfsod}, demonstrate that our method outperforms the approach by Lee et al.~\cite{Lee22Rethink}. This highlights the effectiveness of our method, surpassing their results across the evaluated tasks, further solidifying the potential of our approach in addressing cross-domain few-shot object detection challenges.

\begin{table*}[t]
\small
\centering
\caption{Performance Results (mAP) Mo-FSOD benchmarks    ~\cite{fu2025cross}. The $\dagger$ denotes that the methods are developed or the results are reported by Fu et al.  ; $\ddag$ indicates the results of MM-FSOD are reported by us. Highest scores are in bold font.}
\label{tab:balanced cdfsod}
\resizebox{\textwidth}{!}{
\begin{tabular}{c|c|c|ccc|ccc|ccc}
\toprule
\multirow{2}{*}{\diagbox[height=20pt,innerrightsep=25pt]{\textbf{Method}}{\textbf{Shot}}} & \multirow{2}{*}{\textbf{Backbone}} & \multirow{2}{*}{\textbf{Modality}} & \multicolumn{3}{c|}{\textbf{Clipart1k}} & \multicolumn{3}{c|}{\textbf{DeepFish}} & \multicolumn{3}{c}{\textbf{NEU-DET}} \\
			& & & 1 & 5 & 10 & 1 & 5 & 10 & 1 & 5 & 10 \\
\midrule

ViTDeT-FT$\dagger$ ~\cite{li2022exploring} & ViT-B/14 & Single & 6.1 & 23.3 & 25.6 & 0.9 & 9.0 & 13.5 & 2.4 & 6.5 & 15.8 \\

Detic$\dagger$~\cite{zhou2022detecting} & ViT-L/14 & Multi & 11.4 & 11.4 & 11.4 & 0.9 & 0.9 & 0.9 & 0.0 & 0.0 & 0.0 \\
Detic-FT$\dagger$ ~\cite{zhou2022detecting} & ViT-L/14 & Multi & 15.1 & 20.2 & 22.3 & 9.0 & 14.3 & 17.9 & 3.8 & 14.1 & 16.8 \\
DE-ViT$\dagger$~\cite{zhang2023detect} & ViT-L/14 & Single & 0.5 & 5.5 & 11.0 & 0.4 & 2.5 & 2.1 & 0.4 & 1.5 & 1.8 \\
Meta-DETR$\dagger$~\cite{Zhang23MetaDETR} & ViT-L/14 & Single & 9.6 & 17.3 & 20.7 & 4.1 & 7.5 & 13.6 & 2.0 & 13.4 & 15.2 \\
\midrule 
Next-Chat$\ddag$~\cite{zhang2023nextchat} & ViT-L/14 & Multi & 5.5 & 6.1 & 12.1 & 0.4 & 1.9 & 3.0 & 1.4 & 1.6 & 2.0 \\
\midrule
\makecell{\textbf{Our Method} \\ w/ self-built text} & DETR-R101 & Multi & \textbf{22.9} & \textbf{42.6} & \textbf{47.7} & \textbf{21.2} & \textbf{23.3} & \textbf{25.2} & \textbf{6.1} & \textbf{15.2} & \textbf{19.9} \\
\makecell{\textbf{Our Method} \\ w/ LLM text} & DETR-R101 & Multi & \textbf{25.6} & \textbf{44.3} & \textbf{48.8} & \textbf{23.5} & \textbf{25.6} & \textbf{26.4} & 5.8 & 13.1 & 13.2 \\
\bottomrule
\end{tabular}
}
\end{table*}

\begin{table*}[t]
\small
\centering
\caption{Performance Results (mAP) on CDS-FSOD benchmark   ~\cite{Lee22Rethink}. Highest scores are in bold font.}
\label{tab:unbalanced cdfsod}
\resizebox{ \textwidth}{!}{
\begin{tabular}{c|c|c|cc|cc}
\toprule
\multirow{2}{*}{\diagbox[height=20pt,innerrightsep=25pt]{\textbf{Method}}{\textbf{Shot}}} & \multirow{2}{*}{\textbf{Backbone}} & \multirow{2}{*}{\textbf{Modality}} & \multicolumn{2}{c|}{\textbf{Clipart}} & \multicolumn{2}{c}{\textbf{SIXray}} \\
			& & & 1 & 5 & 1 & 5 \\
\midrule
Lee et al.   & ResNet-50 & Single & 37.2 & 49.3 & 6.6 & 23.9 \\
\midrule
\makecell{\textbf{Our Method} \\ w/ self-built text} & DETR-VIT-L & Multi & \textbf{46.5} & \textbf{58.4} & \textbf{13.3} & \textbf{34.7} \\
\makecell{\textbf{Our Method} \\ w/ LLM text} & DETR-VIT-L & Multi & \textbf{47.2} & \textbf{58.8} & \textbf{14.5} & \textbf{35.3} \\
\bottomrule
\end{tabular}
}
\end{table*}

Across all benchmarks, we observe that when rich text descriptions of the target objects are available, few-shot learning performance improves significantly. However, the quality and context of the rich text are crucial for effective knowledge transfer. Users must carefully assess and fine-tune the rich text for optimal results. This process may pose challenges in two situations: (i) when the rich text includes irrelevant descriptions, or (ii) when the user lacks rich text and needs to generate it from scratch. In the first case, focusing on descriptions most relevant to the visual features of the target objects is recommended. In the second case, users could enhance and refine LLM-generated text descriptions to improve their effectiveness.

\subsection{Analytic Experiments}
\label{sec: ablation}

We offer analytic experiments to offer a better insight about our proposed approach.

\subsubsection{Effect of the Rich text length}

We first investigate the impact of the rich text length on model performance by conducting experiments on the UODD dataset. We compare four text cases: (i) the raw category name, (ii) our manually-built rich text referencing Wikipedia, (iii) an extended version of (ii) with added image context, and (iv) LLM-generated rich text. Cases (ii) to (iv) involve longer, semantically-rich sentences, generally improving model performance, as shown in Table~\ref{tab:text length}. However, not all sentence length extensions are beneficial. For instance, in case (iii), extending the description for sea cucumbers slightly reduced performance from 17.4 to 16.9, likely due to confusion caused by the inclusion of sea urchin, another category with distinct characteristics in the dataset.

\begin{table*}[ht]
    \centering
	\resizebox{1\textwidth}{!}{
    \begin{tabularx}{\textwidth}{l|X|c}
        \toprule
        Type of text & Example & UODD \\
        \midrule
        None & e.g., None &  9.2 \\
        \midrule
        Category name & e.g., Sea cucumbers &  9.9 \\
        \midrule
        Rich text & e.g., Sea cucumbers have sausage-shape, usually resemble caterpillars; their mouth is surrounded by tentacles & 17.4 \\
        \midrule
        Extended rich text & e.g., Sea cucumbers have sausage-shape, usually resemble caterpillars; their mouth is surrounded by tentacles; usually seen together with sea urchins. & 16.9 \\
        \midrule
        LLM-Rich text & e.g., Sea cucumbers are marine invertebrates known for their elongated, leathery bodies, which are typically covered in spines or tentacles and lack a distinct head or tail. & 17.5 \\
        \bottomrule
    \end{tabularx}
    }
    \caption{The effect of the length of language modalities on the final performance. We provide the performance when four types of text descriptions are used for the classes.}
    \label{tab:text length}
\end{table*}

\subsubsection{Module Analysis}

We evaluate the effectiveness of our proposed multi-modal aggregated feature module and the rich semantic rectify module. As shown in Table~\ref{tab:modules}, our method significantly outperforms the single-modal baseline by incorporating rich text multi-modal information. Additionally, the rich text rectify module further enhances performance, demonstrating its effectiveness in improving the overall model accuracy. This indicates that leveraging both rich text and the rectify module plays a critical role in boosting the model's ability to learn from multi-modal data.

\subsubsection{Effect of Information Modalities  on Performance} 

We perform an ablation experiment, as shown in Table~\ref{tab:modality}, to assess the impact of different modalities on downstream performance. When only the image modality is used, the model performs at the baseline level. Introducing the language modality, which uses an encoder language branch to extract prototype features, leads to a performance drop. This is because the language modality is tailored to categories rather than individual images, limiting its effectiveness for task-level support. However, as demonstrated in Table~\ref{tab:modules}, combining the language prototype with vision-language alignment significantly improves performance.

\subsubsection{Effect of Shared vs Decoupled Attention} 
In our MM Aggregation module, we employ a shared attention mechanism. To assess the effectiveness of this choice, we also compare it with a decoupled attention mechanism, where multi-head self-attention modules are separated for the support vision and semantic prototypes. As shown in Table~\ref{tab:decoupled attention}, the decoupled attention approach results in a decrease in performance. This decline occurs because the language prototypes are designed at the category level, not the image level. Thus, using separate attention modules diminishes the benefits of the language modality for task-level support.

\begin{table}[t]
    \centering
	\resizebox{0.8\linewidth}{!}{
    \begin{tabular}{l|c|c|c}
        \toprule
        Module & ArTaxOr & DIOR & UODD \\
        \midrule
        Meta-DETR & 14.5 & 19.9 & 9.2 \\
        Meta-DETR + MM Aggre. & 59.8 & 30.1 & 15.7 \\
        Meta-DETR + MM Aggre. + Rich Text Rect. & 61.4 & 31.4 & 17.5 \\
        \bottomrule
    \end{tabular}
    }
    \caption{The effect of different modules on the final performance.}
    \label{tab:modules}
\end{table}

\begin{table}
\small
    \centering
    \resizebox{0.65\linewidth}{!}{
    \begin{tabular}{c|c|c|c}
        \toprule
        Method & ArTaxOr & DIOR & UODD \\
        \midrule
        Image modality only & 14.5 & 19.9 & 9.2 \\
        Language modality only & 11.3 & 7.7 & 5.1 \\
        \bottomrule
    \end{tabular}
    }
    \caption{\small Ablation experiments using single modalities. In each experiments, one modality is occluded to study its effect on the final performance.}
    \label{tab:modality}
\end{table}

\begin{table}
\small
    \centering
    \resizebox{0.6\linewidth}{!}{
    \begin{tabular}{c|c|c|c}
        \toprule
        Method & ArTaxOr & DIOR & UODD \\
        \midrule
        Shared attention & 61.4 & 31.4 & 17.5 \\
        Decoupled attention & 46.1 & 17.7 & 12.4 \\
        \bottomrule
    \end{tabular}
    }
    \caption{The effect of using shared vs decoupled attention modules on the final performance.}
    \label{tab:decoupled attention}
\end{table}

\begin{figure}[t]
    \centering
    \includegraphics[width=\linewidth]{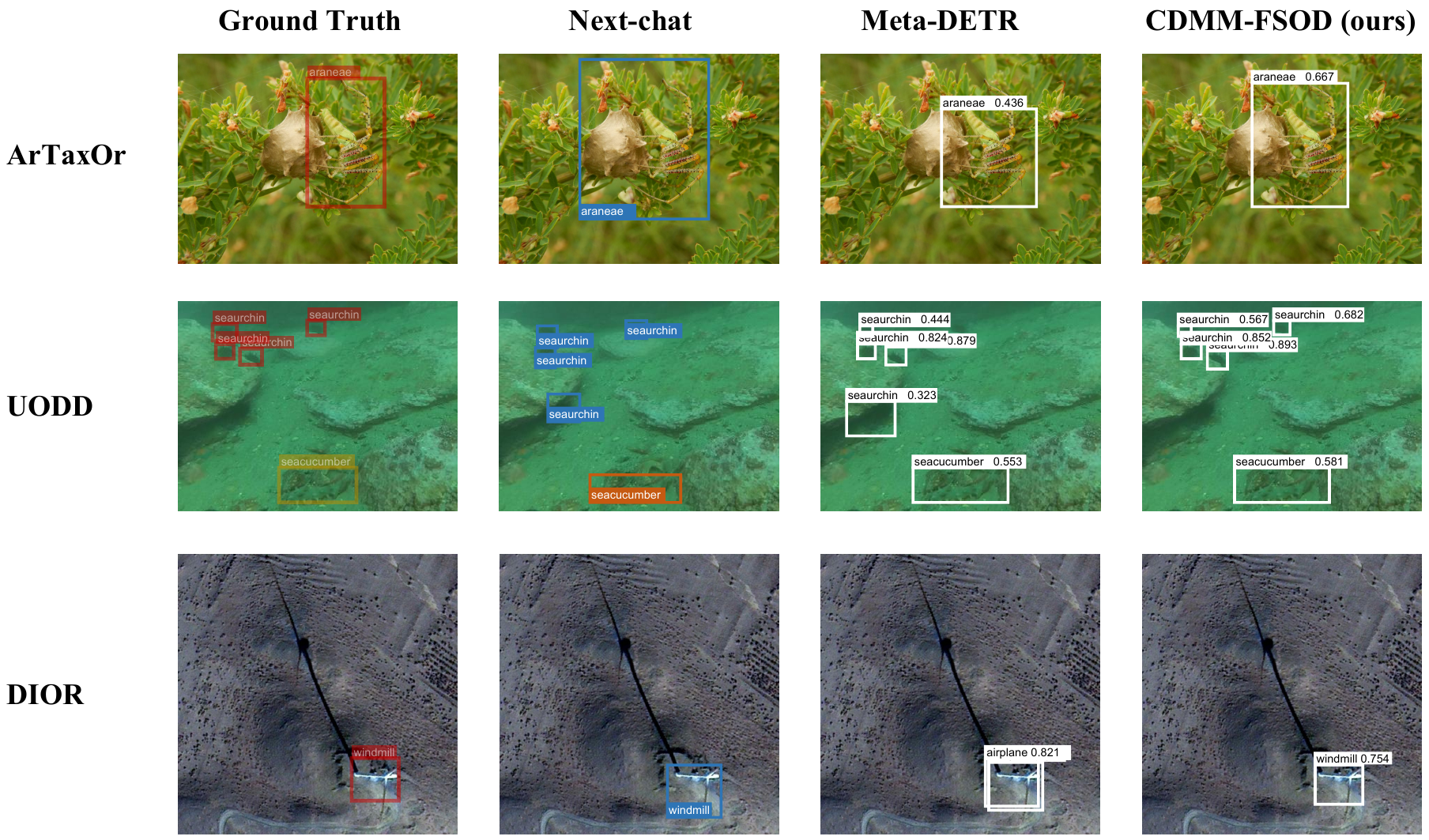}
\caption{Representative visualizations of detection results on three benchmark datasets. We compare our method with a multi-modal object detection model (Next-chat) and a few-shot object detection model (Meta-DETR). Our proposed model obtains more accurate bounding boxes and improved detection confidence.}
    \label{fig:visualization}
\end{figure}

\subsection{Qualitative Comparison}
\label{sec:visualization}

In this section, we present qualitative detection results to illustrate our method's effectiveness. Figure~\ref{fig:visualization} showcases examples from the three CD-FSOD benchmarks, where the goal is to detect araneae (ArTaxOr), sea urchins and sea cucumbers (UODD), and windmills (DIOR). Using a confidence threshold of 0.3 for all methods, we observe improvements over baseline methods. These include more precise bounding boxes (first row), fewer false positives (second row), and higher detection confidence (third row).

\begin{figure*}[t]
    \centering
    \includegraphics[width=\linewidth]{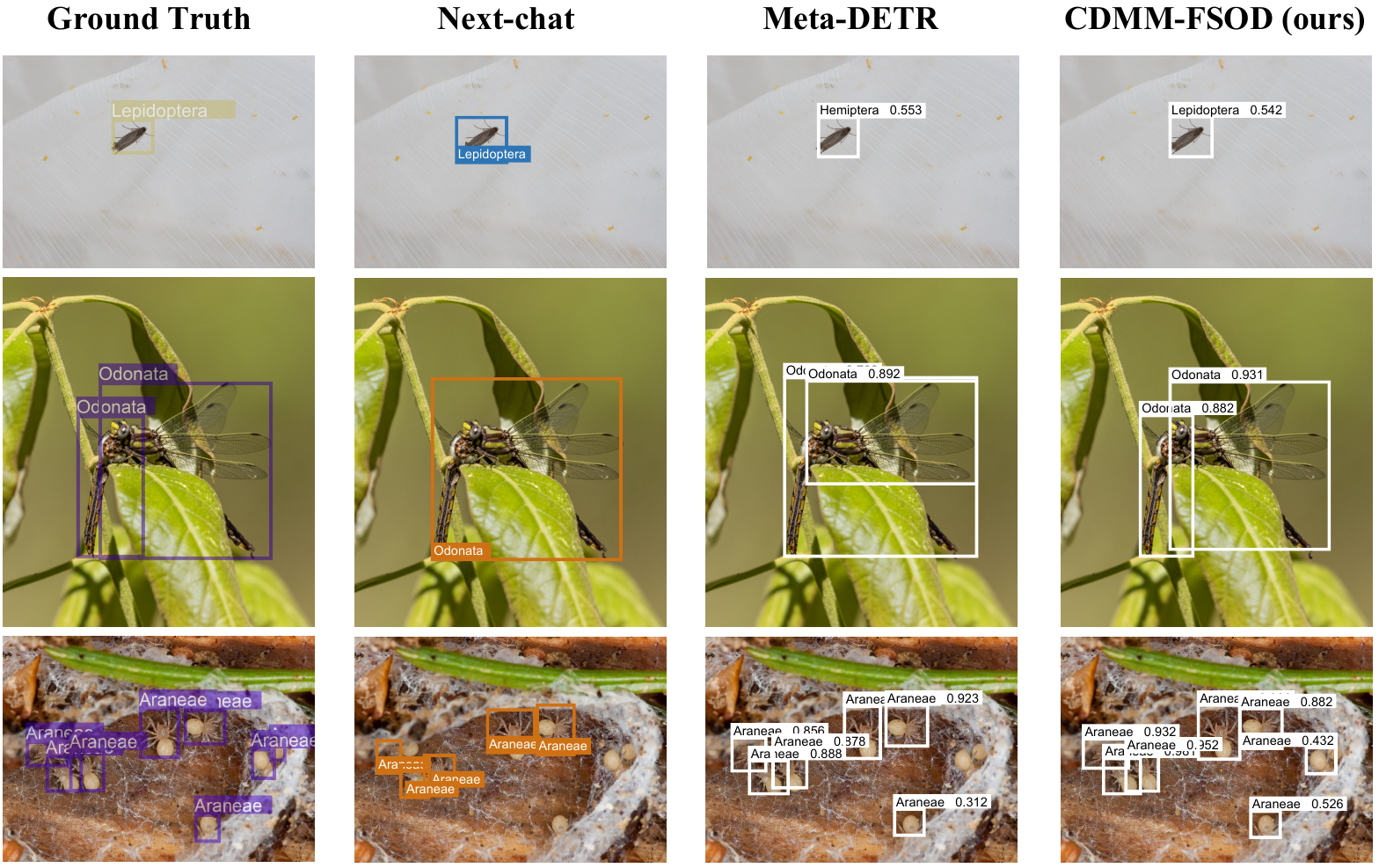}
    \caption{Examples of detection results from ArTaxOr.}
    \label{fig:ArTaxOrPlus}
\end{figure*}

\begin{figure*}[t]
    \centering
    \includegraphics[width=\linewidth]{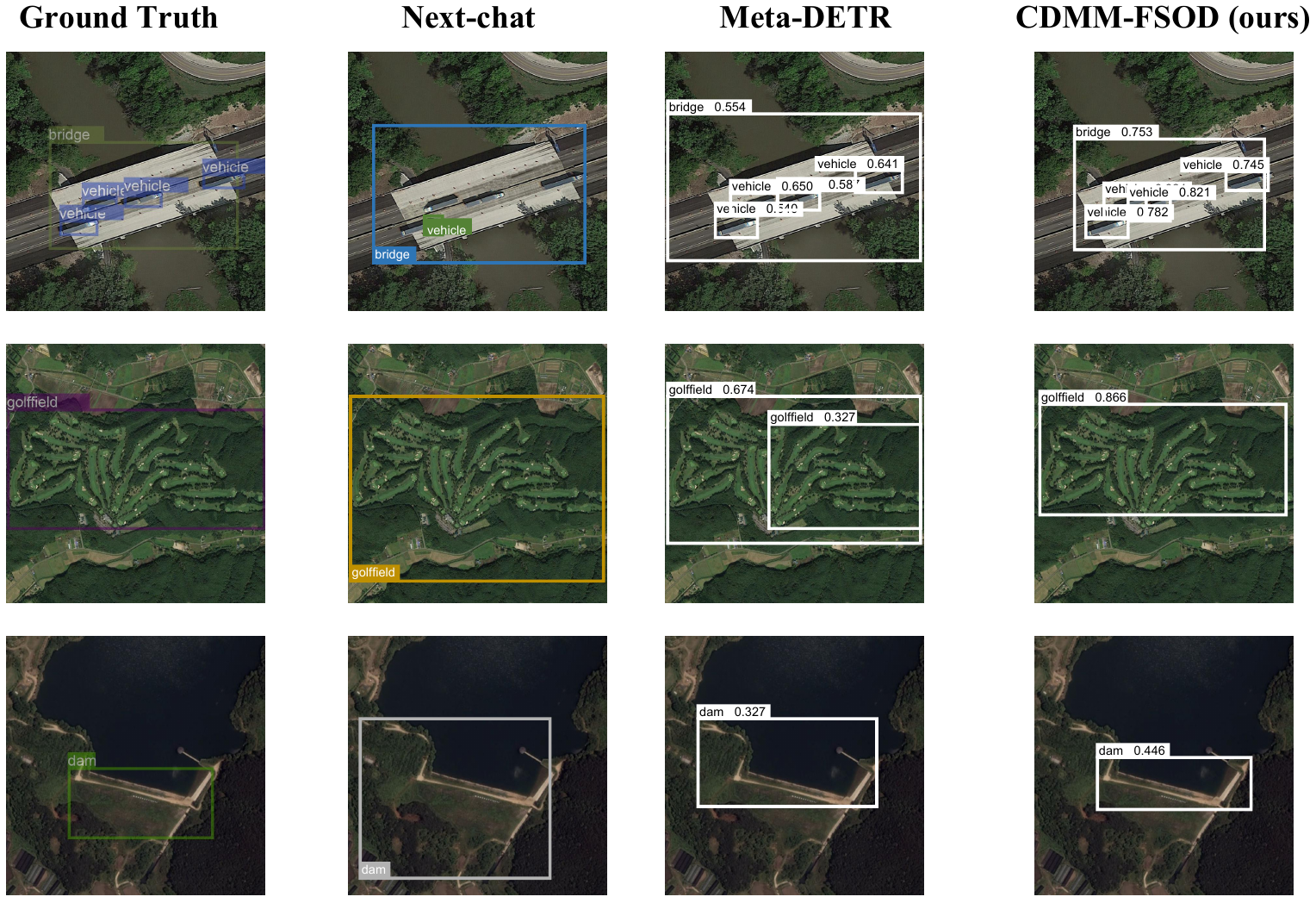}
    \caption{Examples of detection results from DIOR.}
    \label{fig:DIORPlus}
\end{figure*}

\begin{figure*}[t]
    \centering
    \includegraphics[width=\linewidth]{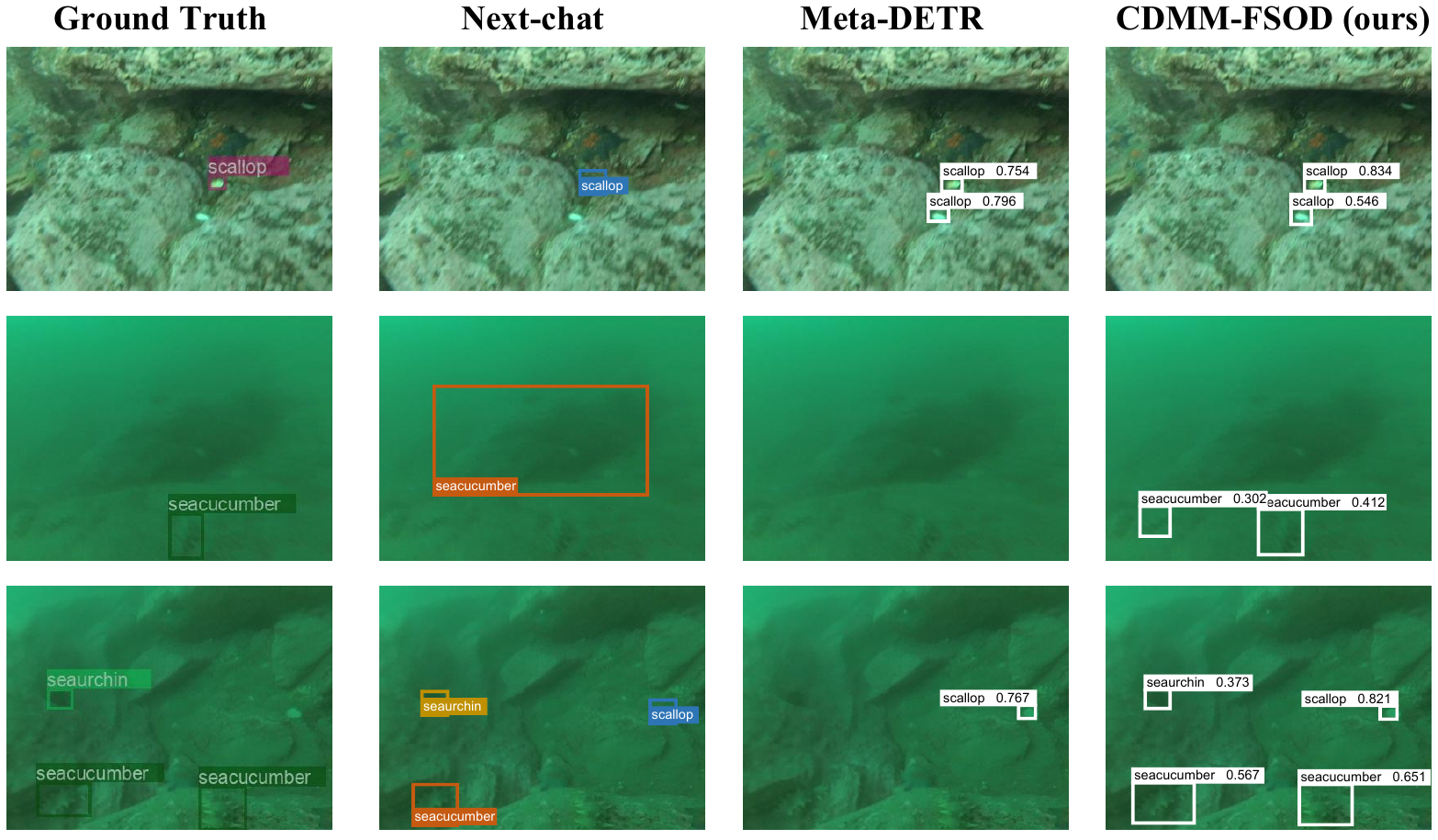}
    \caption{Examples of detection results from UODD.}
    \label{fig:UODDPlus}
\end{figure*}

 The detection results for ArTaxOr, DIOR, and UODD are shown in Figures~\ref{fig:ArTaxOrPlus}, \ref{fig:DIORPlus}, and \ref{fig:UODDPlus}, respectively. Consistent improvements are observed across all datasets when compared to both the multi-modal method (Next-chat) and the single-modal method (Meta-DETR). For instance, in Figure~\ref{fig:ArTaxOrPlus}, there is better classification of Lepidoptera, improved bounding box regression for Odonate, and stronger classification confidence. Similar improvements are seen in Figures~\ref{fig:DIORPlus} and \ref{fig:UODDPlus}, where additional foreground detections, like the sea cucumber and scallop, are also identified.

\subsection{Computational Overhead}

We provide a reference for our computational overhead in Table~\ref{tab:parameters}, with values based on 10-shot (68 images per epoch) experiments on the ArTaxOr dataset. Our method involves more than 10 times the number of training parameters compared to Meta-DETR, yet requires only about twice the training time. This is due to the calculation and caching of feature embeddings for the rich text. We conclude that the additional computational cost is reasonable, given the substantial performance improvements observed in our experiments.

\begin{table*}[t]
\centering
\begin{tabular}{l|l|l|l}
    \toprule
    & Trainable parameters & Model size & Training time \\
    \midrule
    Meta-DETR & 11,649,794 & 317MB & 36 s/epoch \\
    Our method & 162,927,107 & 671MB & 84 s/epoch \\
    \bottomrule
\end{tabular}
\caption{Computation overhead of our methods}
\label{tab:parameters}
\end{table*}

\section{Conclusion}
We proposed a novel cross-domain multi-modal few-shot object detection network and demonstrated that incorporating rich text information significantly enhances detection robustness in out-of-domain, few-shot settings. Our experiments highlight that the design of rich text, particularly accurate descriptions of object appearance, plays a crucial role in improving performance. Our method outperforms existing approaches on four cross-domain datasets. We hope this work encourages further research into leveraging multi-modality to bridge domain gaps in other computer vision tasks.

\section*{Data Availibility}

The  databases used in the manuscript are publicly available.

{ 
\bibliography{ref}
}

\end{document}